\documentclass[lettersize,journal]{IEEEtran}
\usepackage{amsmath,amsfonts}
\usepackage{algorithmic}
\usepackage{array}
\usepackage{textcomp}
\usepackage{stfloats}
\usepackage{url}
\usepackage{verbatim}
\usepackage{graphicx}
\usepackage{cite}
\usepackage{multirow}
\usepackage{booktabs}
\usepackage{diagbox}
\usepackage[colorlinks,
            linkcolor=blue,
            anchorcolor=blue,
            citecolor=red]{hyperref}
\usepackage{subfigure}

\usepackage{epsfig} 
\usepackage{times} 
\usepackage{amssymb}  
\usepackage{graphics}
\usepackage{color}
\usepackage{bigstrut}
\usepackage{makecell}
\usepackage[table]{xcolor}
\usepackage{rotating}
\usepackage{bbding}
\usepackage[font=small]{caption}
\usepackage{balance}

\newcommand{\revise}[1]{\textcolor{black}{{#1}}}
\newcommand{\revisetwo}[1]{\textcolor{black}{{#1}}}
 \usepackage{array}

\let\oldtwocolumn\twocolumn
\renewcommand\twocolumn[1][]{
     \oldtwocolumn[{#1}{
     \begin{center}
     \centering
     \includegraphics[width=0.86\textwidth]{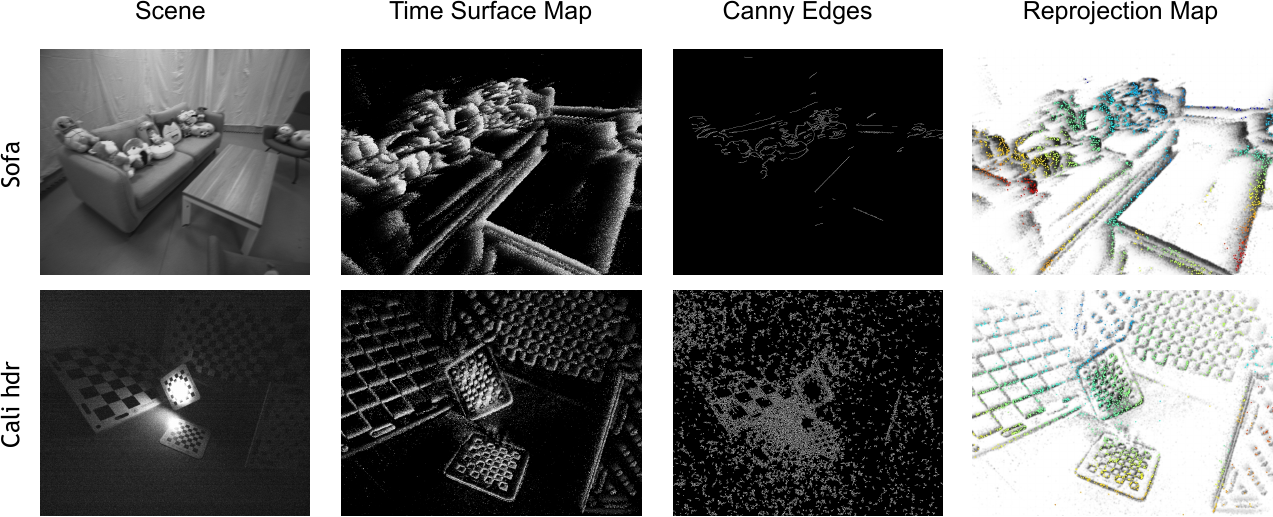}
     \captionof{figure}{\emph{Challenging scenarios and results}. 
	 \textit{sofa} is captured under high dynamics. \textit{cali hdr} is a challenging illumination scene with highly self-similar texture. Column 1: Example RGB images. Column 2: Corresponding time-surface map. Column 3: Canny edge detections. Column 4: Reprojected and aligned semi-dense depth map.}
\end{center}
 }]
}

\begin{document}

\title{Cross-Modal Semi-Dense 6-DoF Tracking of an Event Camera in Challenging Conditions}

\author{Yi-Fan Zuo$^{1}$,
Wanting Xu$^{2}$,
Xia Wang$^{1}$,
Yifu Wang$^{2\dagger}$,
and Laurent Kneip$^{2\dagger}$,
        
\thanks{$^{\dagger}$ indicates corresponding author}
\thanks{$^{1}$ Key Laboratory of Optoelectronic Imaging Technology and Systems, Ministry of Education, School of Optics and Photonics, Beijing Institute of Technology, Beijing 100081, China.
$^{2}$Mobile Perception Lab, ShanghaiTech University.} 
}

\maketitle

\begin{abstract}
Vision-based localization is a cost-effective and thus attractive solution for many intelligent mobile platforms. However, its accuracy and especially robustness still suffer from low illumination conditions, illumination changes, and aggressive motion. Event-based cameras are bio-inspired visual sensors that perform well in HDR conditions and have high temporal resolution, and thus provide an interesting alternative in such challenging scenarios. \revise{While purely event-based solutions currently do not yet produce satisfying mapping results, the present work demonstrates the feasibility of purely event-based tracking if an alternative sensor is permitted for mapping. The method relies on geometric 3D-2D registration of semi-dense maps and events, and achieves highly reliable and accurate cross-modal tracking results. Practically relevant scenarios are given by depth camera-supported tracking or map-based localization with a semi-dense map prior created by a regular image-based visual SLAM or structure-from-motion system.} Conventional edge-based 3D-2D alignment is extended by a novel polarity-aware registration that makes use of signed time-surface maps (STSM) obtained from event streams. We furthermore introduce a novel culling strategy for occluded points. Both modifications increase the speed of the tracker and its robustness against occlusions or large view-point variations. The approach is validated on many real datasets covering the above-mentioned challenging conditions, and compared against similar solutions realised with regular cameras.
\end{abstract}


\section{Introduction}

\IEEEPARstart{R}{eal-time} localization and tracking are increasingly important tasks to be solved in many emerging technologies such as robotics, intelligent transportation, and intelligence augmentation. To achieve highly robust and accurate localization, most intelligent mobile devices are equipped with multiple sensors, such as cameras, lidars, depth cameras, Inertial Measurement Units (IMU), or GPS. Owing to their small scale and affordability, cameras are often considered as the preferred choice for exteroceptive sensing. However, despite the fact that a certain level of maturity has already been reached, pure vision-based solutions lack robustness in situations of high dynamics, low texture distinctiveness, or challenging illumination conditions~\cite{fuentes2015visual,cadena2016past}.

Event cameras---also called dynamic vision sensors---represent an interesting alternative visual sensor that pairs HDR with high temporal resolution. The potential advantages and challenges behind event-based vision are well explained by the original work of Brandli~et~al.~\cite{brandli2014240} as well as the recent survey by Gallego~et~al.~\cite{gallego2019event}. A number of solutions for purely event-based~\cite{gallego2017event,kim2008simultaneous,weikersdorfer2013simultaneous,reinbacher2017real,zhou2021event,rebecq2016evo,kim2016real} or hybrid~\cite{censi2014low,weikersdorfer2014event,kueng2016low,zuo2022devo,vidal2018ultimate,rebecq2017real,zihao2017event,mueggler2018continuous,le2020idol} tracking and mapping have already been presented in the literature. However, given that event cameras do not output absolute intensities but rather intensity changes, photometric-consistency-based methods cannot be directly applied, and feature extraction tends to be unstable. Besides, limited spatial resolution and low signal-to-noise ratio pose challenges to the quality of the 3D map built from events, which is why event-based mapping and registration often struggles to achieve the same level of accuracy than standard cameras~\cite{engel2017direct,engel2014lsd,forster2014svo,mur2015orb}. The community therefore has put substantial effort into the development of methods that use image-based maps as a reference for event-based tracking~\cite{8094962, bryner2019event}. 
\revise{The interest in cross-modal event-based tracking is further supported by the fact that many scenarios permit the use of other sensors towards mapping. Examples are given by systems that are equipped with a consumer depth camera, thereby providing a reliable way to retrieve local scene depth. Another example is given by prior map-based tracking, where the reference map is generated a priori through the use of images. It is important to note that the use of alternative sensors for mapping does not necessarily undermine the potential advantages given by event cameras. Depth cameras work well in low illumination, while illumination conditions or dynamics during the dedicated generation of map priors can be controlled.}

\revise{In this work, we present computationally efficient, semi-dense cross-modal 6-dof tracking of a single event camera aiming specifically at such scenarios. The semi-dense 3D map of the environment is either built locally by leveraging information from a depth camera, or globally by a prior run of a monocular visual SLAM or structure-from-motion framework. The corresponding tracking methods are denoted \textit{Canny-DEVO (Depth-Event Visual Odometry} and \textit{Canny-EVT (Event-based Visual Tracking)}. Both methods proceed by extracting edge maps from the event stream and registering the camera poses by semi-dense, 3D-2D edge alignment. The properties of event cameras lead to excellent performance even in challenging cases while maintaining high computational and energy efficiency. We believe that proposed solution has wide applicability in scenarios in which a depth camera or an open mapis available (e.g. AR, Autonomous Parking or Campus Navigation).}
Our detailed contributions are as follows:

\begin{itemize}
    \item We present a novel cross-modal tracking approach for event cameras. The method relies on semi-dense 3D point cloud priors obtained from reliable depth signals or regular camera-based mapping algorithms. Events are only used for tracking, which enables highly accurate results even under challenging conditions.
    \item We use Signed Time Surface Maps (STSMs) to split the registration cost into two individual energies and help to increase the convergence basin and improve accuracy in highly dynamic situations. We furthermore introduce a novel preemptive semi-dense point registration strategy that discards occluded points if their potential match is registered to a more reasonable foreground point.
    \item We release the code for our framework\footnote{\href{https://github.com/zyfff/Canny-EVT/}{https://github.com/zyfff/Canny-EVT/}}. Besides the \textit{Canny-DEVO} and \textit{Canny-EVT}, the framework supports regular camera alternatives (denoted \textit{Canny-VO}~\cite{zhou2018canny} and \textit{Canny-VT}, respectively).
\end{itemize}

Our results are obtained on both self-collected and publicly available high-resolution RGB-D-event indoor datasets with ground truth captured by an external motion tracking system. Further tests are conducted on larger scale outdoor datasets where depth is obtained from a LiDAR scanner. We thoroughly compare the event-based cross-modal trackers against vision-based alternatives, state-of-the-art RGB-D odometry~\cite{newcombe2011kinectfusion}, as well as event-based methods that rely on purely event-based mapping~\cite{zhou2021event}. Our comparison demonstrates that semi-dense cross-modal tracking achieves highly accurate and efficient, continuous visual localization results that eventually outperform alternative methods under challenging conditions.


The paper is organized as follows. Section \ref{sec:relatedWork} introduces further related work. Section \ref{sec:preliminaries} provides preliminaries including an overview of the frameworks introduced and published open-source through this work. Section \ref{sec:semidense_mapping} explains both local and global mapping, while Section \ref{sec:6dof_tracking} provides all details on our semi-dense tracking approach. Section \ref{sec:experiments} presents all experimental results, while Section \ref{sec:conclusion} concludes the work. Note that the present paper is an extension of our previous work~\cite{zuo2022devo}.

\section{Related work}\label{sec:relatedWork}

There is a large number of successful works related to vision-based localization over the past two decades. The present literature review focuses on works that are closely related to our proposed event-based tracking with semi-dense maps.

\textbf{RGB and stereo-based VO, SLAM, and tracking:} MonoSLAM~\cite{davison2007monoslam} is the first monocular SLAM system and realized with an extended Kalman filter back-end. Later on, PTAM~\cite{klein2007parallel} represents the first keyframe-based SLAM system that divides tracking and mapping into two parallel threads. PTAM shows better performance than filter-based systems owing to iterative bundle adjustment. Its successor---ORB-SLAM~\cite{davison2007monoslam,mur2017orb,campos2021orb3}---is also a keyframe-based SLAM system but distinguishes itself by uniform use of the efficient binary ORB features for tracking, mapping, and loop detection. The system furthermore includes an extra thread for loop detection and closure. Contrasting with feature-based methods, recent years have seen the verge of direct methods as a valid and efficient alternative for real-time SLAM systems. These methods do not extract features but directly minimize photometric errors. One of the first systems to perform direct estimation is the semi-dense visual odometry (SDVO) approach by Engel et al.~\cite{engel2013semi}, which proposes semi-dense epipolar tracking and incremental depth filtering to make the system run in real-time on a CPU. Later on, the same authors propose LSD-SLAM~\cite{engel2014lsd} which builds on top of SDVO but extends the method by adding loop-closure and large-scale semi-dense mapping. In order to further reduce computational burden, SVO~\cite{forster2014svo,forster2016svo} uses a direct method to track sparse features. Direct Sparse Odometry (DSO)~\cite{engel2017direct} is based on local photometic bundle adjustment optimization and improves performance in low textured environments. The literature on the topic is vaste, and the interested reader is kindly referred to the survey by Cadena et al.~\cite{cadena2016past} for further reading. \revise{Note that the present review focuses on traditional methods and does not cover more modern approaches that rely on deep representations in either front-end~\cite{Min_2020_CVPR} or back-end~\cite{lin2021barf, rosinol2022nerfslam, NEURIPS2021_89fcd07f}.}
 
\textbf{RGB-D camera-based tracking:}
The most straightforward solutions to RGB-D camera-based VO use only depth information \cite{newcombe2011kinectfusion,whelan2016elasticfusion}. While they may potentially operate in dark environments, they require dense depth image processing at high frame rate, and therefore require high energy and computation resources (e.g. GPU). Approaches that also rely on images~\cite{steinbrucker2011real, endres2013rgbd, kerl2013robust, henry2014rgb} often perform dense photometric alignment, and thus still depend on exhaustive parallel computing. They furthermore have the disadvantage of degrading in challenging visual conditions (e.g. blur, low illumination). Most related to our method are approaches relying on sparsified, semi-dense depth maps \cite{zhou2018canny, schenk2019reslam}. They have large convergence basins, stability under illumination changes, and high computational efficiency. Nonetheless, they still depend on intensity images for edge detection, and therefore continue to demonstrate high sensitivity to motion blur and low-illumination conditions. Note that the method of Zhou et al.~\cite{zhou2018canny} serves as a basis for the event-based methods presented in this work, and an implementation will be contained in our open-source release.

\textbf{Pure event camera-based mapping and tracking:} Event cameras offer strong advantages such as high dynamic range, low latency, and low power consumption. However, the complicated nature of event data demands novel theories and approaches. Full 6-DoF motion estimation with a single event camera remains a challenging problem. Many works rely on simplifying assumptions. Weikersdorfer et al.~\cite{weikersdorfer2013simultaneous} proposed an event-based 2D SLAM framework for planar motions. Other works rely on a contrast maximization objective that utilizes image-to-image warping, a function that only works if the image transformation is at most a homography (e.g. pure rotation, planar homography)~\cite{gallego2017accurate, stoffregen2019event, gallego2019focus, liu2020globally, peng2021globally}. The first full 6-DoF solution is given by Kim et al.~\cite{kim2016real}, who proposed a complex framework of three decoupled probabilistic filters estimating intensity, depth, and pose, respectively. A geometric solution is given by Rebecq et al.~\cite{rebecq2016evo}, which relies on their earlier ray-density based structure extraction method EMVS~\cite{rebecq2018emvs}. Wang et al.~\cite{wang2022visual} proposed a similar approach but jointly optimize structure and motion. The success of the listed methods is limited to small-scale environments and small, dedicated movements to ensure healthy map information. \revise{Chamorro et al.~\cite{9810191} propose a line feature-based SLAM method which minimizes event-line reprojection errors using an error-state Kalman filter. The method however depends on the sufficient availability of straight line features.} Zhu et al.~\cite{zhu2019neuromorphic} finally present a promising learning-based approach, which however depends on vast amounts of training data, and provides no guarantees of optimality or generality. ESVO leverages information from a stereo event camera~\cite{zhou2021event}, \revise{which is similar to our tracking module but does not utilize the polarity of events and neither handles occluded points.} Note that we compare our method against their approach. It is important to note that none of the listed methods can achieve the same level of accuracy as state-of-the-art standard camera-based solutions (e.g.~\cite{campos2021orb3}, \cite{forster2016svo} and \cite{engel2017direct}). Generally speaking, the map information obtained from events alone is often of inferior quality, which motivates our cross-modal tracking approach.

\textbf{Hybrid event-supported solutions:} Owing to their difficult nature, events are often combined with other sensor data such as IMU readings or regular images. Zhu et al.~\cite{zihao2017event} fused feature tracks with IMU readings in an Extended Kalman Filter to get better performance. Both Le Gentil et al. ~\cite{le2020idol} and Mueggler et al. ~\cite{mueggler2018continuous} leverage continuous-time representations to perform batch optimization fusing events and inertial readings. Censi and Scaramuzza~\cite{censi2014low} present a VO framework that estimates relative poses by fusing events with absolute brightness information. Kueng et al. \cite{kueng2016low} detect features from grayscale images and track the features using the support of event data. \revise{~\cite{hidalgocarrió2022eventaided} also merge events and regular frames in a direct visual odometry framework. The events are leveraged to track the camera motion between successive frames.} Methods that rely on intensity images at their core do not take full advantage of event cameras and may easily fail due to motion blur. On the other hand, approaches that process images and events individually~\cite{rebecq2017real, vidal2018ultimate, mahlknecht2022exploring} may continue to work if no regular image features are perceived, but they often lead to inaccurate local event-based maps and severe robustness issues if degraded visual conditions persist over extended time intervals. Another work that is closely related to ours is introduced by Weikersdorfer et al.~\cite{weikersdorfer2014event}, who extend their previous work~\cite{weikersdorfer2013simultaneous} and include an RGB-D sensor. However, their method is based on an outdated, low-resolution event camera model and relies on a fully voxelized and thus limited-size environment. Gallego et al.~\cite{8094962} and Bryner et al.~\cite{bryner2019event} already present cross-modal event-based tracking. They leverage a dense photometric 3D map to predict events and estimate the 6-dof pose by event camera. \revise{The core idea consists of predicting intensity changes from the photometric 3D map and registering them against the measured events. The method operates densely, and thus induces significant computational burden.} To the best of our knowledge, our work is the first to propose cross-modal event-based tracking based on an efficient geometric, semi-dense paradigm.

\section{Preliminaries}\label{sec:preliminaries}

This section presents an overview of our newly proposed, event-based semi-dense tracking frameworks, followed by details of the original and modified event data representations needed for efficient processing.

\subsection{Framework overview}\label{sec:framework}

\begin{figure}[b!]
	\centering
	\includegraphics[width=1.0\linewidth]{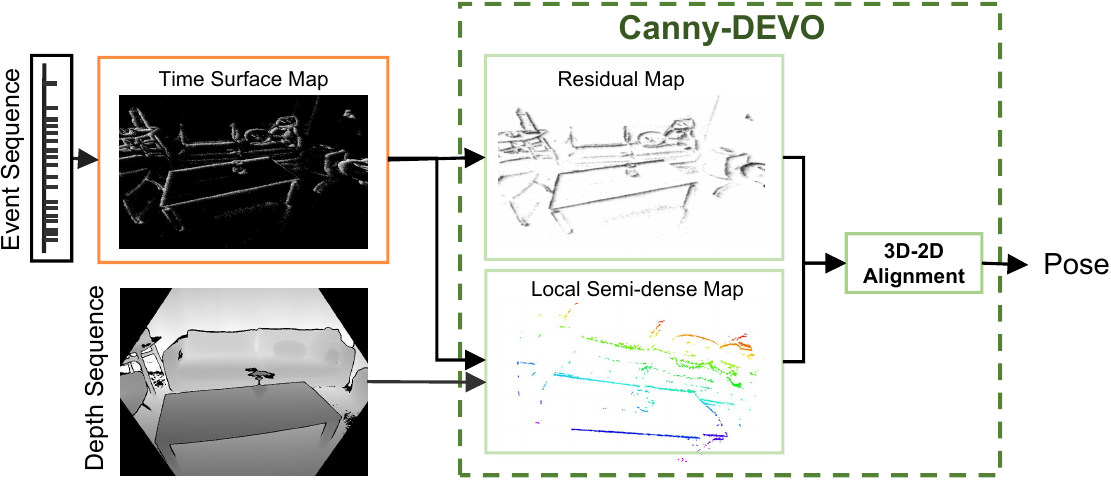}
	\caption{\textit{Overview of our proposed} \textbf{Canny-DEVO} \textit{visual odometry pipeline.}}
	\label{fig:flowchart1}
	\vspace{-0.4cm}
\end{figure}

\begin{figure}[t!]
	\centering
	\includegraphics[width=1.0\linewidth]{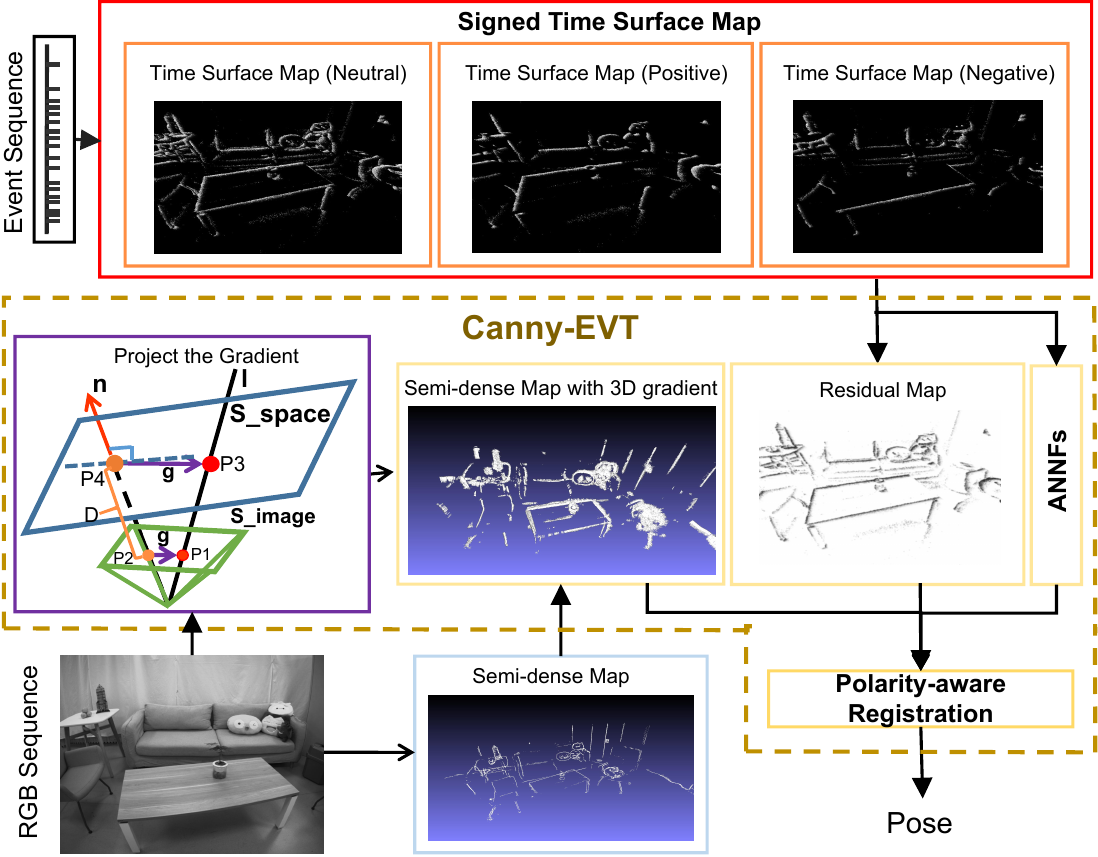}
	\caption{\textit{Overview of our proposed} \textbf{Canny-EVT} \textit{visual tracking pipeline.}}
	\label{fig:flowchart2}
	\vspace{-0.4cm}
\end{figure}

Flowcharts of our proposed methods detailing all sub-modules are illustrated in Figures \ref{fig:flowchart1} and \ref{fig:flowchart2}. We start by generating time-surface maps which put our event sets into a suitable representation for efficient and accurate edge extraction and alignment. Details are introduced in Section \ref{subsubsec:tsm}. We furthermore propose signed time-surface maps (\textbf{STSM}, cf. Section \ref{subsubsec:stsm}), which exploit the polarity of events to split the time surface map into three sub-maps. The use of signed maps ensures compatibility between predicted and measured event polarities, and thereby enlarges the convergence basin during registration.

For \textit{Canny-DEVO}, the representation is used in both the mapping and the tracking module. The mapping module takes the semi-dense edge map and assigns depth values from the depth camera readings, thereby creating local reference semi-dense depth maps online for incremental tracking. The local map is updated by readings from a new depth frame whenever sufficient displacement between the current and the reference view has been detected. The operations of the map generation and the reference frame selection strategy are detailed in Section \ref{mapping module for devo}. The tracking thread processes only events and incrementally estimates the 6-DoF camera pose by efficient 3D-2D edge alignment. Details of the tracking thread are introduced in Section \ref{sec:6dof_tracking}.

For \textit{Canny-EVT}, the time-surface map is only used in the tracking module. The semi-dense map in turn is generated globally from RGB images that have been aligned using online structure-from-motion. Section \ref{mapping module for sdevo} introduces the semi-dense map extraction. We furthermore augment every 3D point in the semi-dense map by a 3D gradient vector by projecting the image gradients onto an approximate support plane. This spatial vector is combined with optical flow propagations in order to predict the polarity of any event triggered by a 3D point. As explained in Section \ref{polarity-aware registration by STSMs}, \textit{Canny-EVT} hereby has the option to register 3D points with our polarity-aware signed time-surface maps. To conclude, we introduce an approximate nearest neighbour field (\textbf{ANNF}), which is used to quickly detect collisions when multiple 3D points attempt registration against the same edge point in the image, and thereby helps to avoid the registration of occluded points (cf. Section \ref{occluded points culling by annnfs}).

\subsection{Event representations}
\label{sec:ts_map}

Let us assume that we are given a set of $N$ events $\mathcal{E}=\{e_{k}\}_{k=1}^{N}$ occurring over a certain time interval. Each event $e_k=\{ \mathbf{x}_k, t_k, b_k \}$ is defined by its image location $\mathbf{x}_k=[\begin{matrix} x_k & y_k \end{matrix}]^T$, timestamp $t_k$, and polarity $b_k$. It is common to not process events asynchronously at the very high rate they occur, but aggregate sets of events accumulated during regularly spaced time intervals into one of three possible representations. The first one is given by space-time volumes of events~\cite{lagorce2016hots}, which are often used in conjunction with accurate but more computationally demanding continuous-time motion representations. The second one is given by simply ignoring the temporal nature of the events, and projecting all events along the temporal dimension onto a virtual binary image in which we then perform feature extraction. Though very efficient, the method re-induces motion blur and requires a careful selection of the time interval length. The third representation is given by time-surface maps (\textbf{TSM}~\cite{lagorce2016hots}), which create an interesting balance between accuracy and efficiency. However, \textbf{TSM}s ignore polarity, which causes loss of information. In this paper, we propose the use of signed time surface maps (\textbf{STSM}s), which extend regular \textbf{TSM}s by reinstalling polarity information.

\subsection{Time Surface Maps}
\label{subsubsec:tsm}

A \textbf{TSM} is an image in which the value at each pixel location $\mathbf{x}$ is a function of an exponential decay kernel given by
\begin{equation}
    \mathcal{T}(\mathbf{x},t) = \text{exp}(-\frac{t - t_{last}(\mathbf{x})}{\tau}),
\end{equation}
where $t$ is an arbitrary time, and $t_{last}(\mathbf{x}) \leq t$ is the timestamp of the last event triggered at $\mathbf{x}$. The higher a pixel value, the more recent the local event. $\tau$ denotes the constant decay rate parameter, which requires careful tuning as a function of motion dynamics. A \textbf{TSM} visualizes the history of moving brightness patterns at each pixel location and emphasizes on locations in which motion has been more recent. The values in a \textbf{TSM} are mapped from $[0, 1]$ to $[0, 255]$ for convenient visualization and processing.
In our work, we use a modified \textbf{TSM} in which we only consider pixels with a value above a certain threshold $\delta$. Depending on the module (i.e. tracking, or mapping), other pixels are set to 0 or discarded.

\subsection{Signed Time Surface Maps}
\label{subsubsec:stsm}

One of the key additions in our work is the polarity-aware registration by signed time surface maps. Simply put, we maintain two maps and update the positive map with only positive events, and the negative map with only negative events. Given that the time-surface maps are later on used to approximate distance fields, polarity-aware registration induces the use of multiple distance fields each one with fewer basins of attraction. The result is an increasing convergence radius and thus better convergence behavior in highly dynamic, large baseline scenarios.

\section{Mapping module}\label{sec:semidense_mapping}

Our framework involves one of two mapping modules for semi-dense point cloud extraction. The first one is an online local mapping approach used in \textbf{Canny-DEVO}. It builds local maps from \textbf{TSM}s and depth images (cf. Section \ref{mapping module for devo}). The second mapping module is used in \textbf{Canny-EVT} and builds a global semi-dense map from a single RGB camera (cf. Section \ref{mapping module for sdevo}). The semi-dense map is furthermore enhanced by 3D gradient vectors, which are used for polarity-aware registration (cf. Section \ref{3D gradient vector for SDEVO+ and SDEVO++}).

\subsection{Mapping for \textbf{Canny-DEVO}}
\label{mapping module for devo}

Let $\mathcal{T}_\text{ref}(\cdot)=\mathcal{T}(\cdot,t_\text{ref})$ be the \textbf{TSM} generated from the set of events $\mathcal{E}$ at time $t_\text{ref}$. The semi-dense region $\mathcal{X}^{\text{ref}}$ for which depth values will be extracted is simply given by all pixels for which the value is larger than $\delta$, i.e. $\mathcal{X}^{\text{ref}} = \{\mathbf{x}\text{ s.t.}\mathcal{T}_\text{ref}(\mathbf{x})>\delta$\}. Based on the assumption that events are pre-dominantly triggered by high-gradient edges in the image, a proper balancing of the decay rate $\tau$ and the threshold $\delta$ will prevent the extracted semi-dense regions to remain thin and align tightly with true appearance contours. \revisetwo{The range of the decay rate $\tau$ is about 10e-3 to 50e-3. If the event camera moves very fast, it will trgger too much events. In this case, we need to adjust the parameter to a smaller number to make the time surface map more clear.}

In order to retrieve the depth value for each point in the semi-dense region, we first warp the depth points from the depth camera at time $t_\text{ref}$ to the event camera. The location in the event camera is given by
\begin{equation}
\mathbf{x}^e_k = \pi_{e}(\mathbf{T}_{ed}\cdot D(z^d_k)\cdot \pi^{-1}_{d}(\mathbf{x}^d_k)),
\end{equation}
where $\pi_{e/d}$ and $\pi^{-1}_{e/d}$ represent the known camera-to-image and image-to-camera transformations of the event and the depth camera, respectively. They are defined as mappings from 2D image space to 3D homogeneous space and vice-versa. $D(a)=\text{diag}(a,a,a,1)$ generates a diagonal matrix with elements $a$, $a$, $a$, and $1$ along the diagonal. $\mathbf{T}_{ed}$ is the known $4\times 4$ Euclidean extrinsic transformation matrix from the depth to the event camera. Finally, $\mathbf{x}_k^d$ and $z_k^d$ are a point and its corresponding depth in the depth camera, and $\mathbf{x}_k^e$ is the warped point in the depth frame. The depth $z^e_k$ at the latter point is easily obtained by
\begin{equation}
z^e_k = 
\revise{
\begin{bmatrix}
0 & 0 & 1 & 0
\end{bmatrix}
}
\cdot\mathbf{T}_{ed}\cdot D(z^d_k)\cdot \pi^{-1}_{d}(\mathbf{x}^d_k).
\end{equation}
Note that the warping maps depth values onto sub-pixel locations rather than event camera pixel centers. It may furthermore induce occlusions or leave pixels with unobserved depths. In order to find a unique depth for each pixel, we create an individual list of nearby warped points from the depth image for each pixel in the semi-dense region. The value of the depth is conditionally set if the pixel is surrounded by warped points from the depth image. A simple depth clustering strategy identifies potential foreground points, and the final value is found by simple interpolation and ray intersection. This ensures that the depth of the pixels in the semi-dense region is always corresponding to foreground points and never affected by occlusions, depth measurement errors, or potential misalignments such as small errors in the extrinsic calibration parameters. The points $\mathbf{p}_{\text{ref}}$ that have a valid depth assigned to them are renormalized and multiplied by their depth, which finally results in the set $\mathcal{P}_{\text{ref}}$ for our semi-dense 3D point cloud. Note that---in combination with the reference frame poses identified by the tracking module---multiple local maps could be merged into a global map using classical point cloud fusion techniques. The present work however limits \textbf{Canny-DEVO} to single reference frames.

\subsection{Mapping module for \textbf{Canny-EVT}}
\label{mapping module for sdevo}

\textbf{Canny-EVT}'s mapping module uses the probabilistic semi-dense mapping technique proposed by Mur-Artal et al.~\cite{mur2015probabilistic}. It processes the keyframes generated by the monocular SLAM framework \textbf{ORB-SLAM}~\cite{mur2015orb} and performs 3D reconstruction of the 3D edges that form perceivable appearance contours in the keyframes. Note that these edges do not necessarily need to be geometric edges in 3D, they may still be appearance edges on a smooth support surface. The semi-dense representation is simply given in the form of a point cloud. \textbf{ORB-SLAM} leverages the advantage of feature-based solutions of providing wide baseline matches despite severe illumination and viewpoint changes. The addition of large-scale loop closures and global bundle adjustment results in highly accurate camera pose estimation, which in turn enables decent semi-dense reconstruction from fixed-pose keyframes.

The main idea behind \textbf{Canny-EVT}'s mapping module is thus well-explained in~\cite{mur2015probabilistic}: The algorithm first determines an image-level prior for the depth range by considering the depth of the bundle adjusted 3D points corresponding to correctly measured ORB features. Each pixel in each high gradient region of each keyframe $K_j$ is then tracked within depth-compatible segments of corresponding epipolar lines in $N$ neighbouring keyframes. This procedure will lead to $N$ inverse depth hypotheses for each such pixel. Owing to the large baseline between consecutive keyframes, the epipolar search evaluates not only photometric intensity differences, but also gradient modulo and orientations for finding the best correspondences. Once pixel-level correspondences are retrieved, inverse depths are easily derived by taking into account camera instrinsics and keyframe poses. The uncertainty of the inverse depth hypotheses is finally propagated from intensity noise. All noise assumptions follow gaussian distributions, which easily permits the subsequent fusion of the multiple inverse depth hypotheses. After the initial semi-dense inverse depth map is computed, further outlier removal, smoothing and region growing steps are applied as proposed in the original method by Engel et al. \cite{engel2013semi}. This step improves both the accuracy and the completeness of the reconstruction. To conclude, each pixel that has been assigned an inverse depth value is subjected to a cross-keyframe consistency check by projecting the associated 3D point into neighbouring keyframes. A final gauss-newton optimization step minimizes depth differences between corresponding hypotheses in different frames. For further details of the semi-dense mapping approach, we kindly refer the reader to the detailed original introductions of semi-dense mapping by Engel et al.~\cite{engel2013semi} and Mur-Artal et al.~\cite{mur2015probabilistic}.

\begin{figure}[t!]
    \centering
    \includegraphics[width=0.65\linewidth]{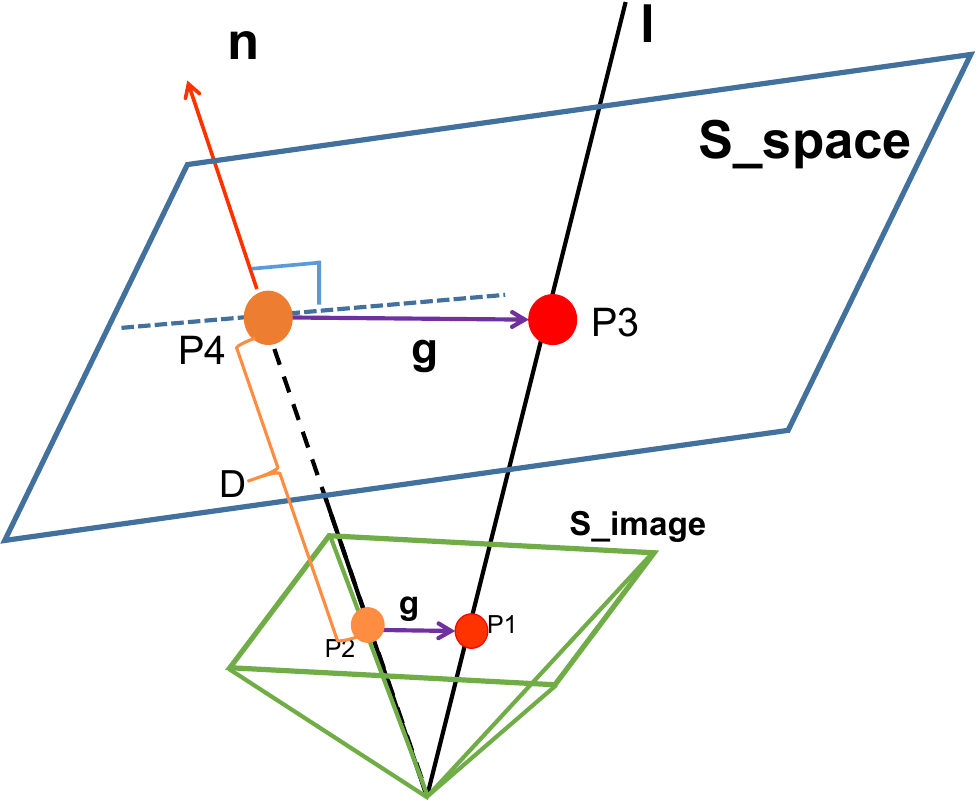}
    \caption{\textit{Projection of the gradient vector}. Projection of the 2D gradient from an RGB image onto the 3D support plane of that point. The normal vector of the support plane is defined to be parallel to the vector pointing from the camera center to the 3D point.}
    \label{fig:project gradient}
    \vspace{-0.3cm}
\end{figure}
\subsection{Addition of 3D gradient vectors}
\label{3D gradient vector for SDEVO+ and SDEVO++}

\textbf{Canny-EVT} can optionally predict the polarity of potential events triggered by a point from the semi-dense map. It does so by predicting the normal flow of the point which in turn is derived from the predicted optical flow and image gradient. The predicted gradient is obtained by projecting 3D gradient vectors into the current frame. These vectors in turn are obtained by projecting the image gradients from the respective observation keyframes onto approximate 3D support planes.

The projection is illustrated in Figure~\ref{fig:project gradient}. We first define the normal vector \textbf{n} of the support plane \textbf{S} of the 3D point \textbf{P4} as being aligned with the vector pointing from the camera center to \textbf{P4}. Picking a second point in the gradient direction in the RGB image and normalizing it to a spatial line \textbf{l}, we can easily find the intersection of that line and the plane \textbf{S}, denoted by the 3D point \textbf{P3}. The vector from \textbf{P4} to \textbf{P3} defines the 3D gradient vector \textbf{g}.

\section{6-Dof camera tracking}
\label{sec:6dof_tracking}

This section defines the tracking problem (cf. Section \ref{sec:tracking problem statement}) and explains how we may use events inside either \textbf{Canny-DEVO} or \textbf{Canny-EVT} to estimate the 6-Dof pose from a semi-dense map (cf. Section \ref{exploting tsm as df}). In Section \ref{polarity-aware registration by STSMs}, we furthermore detail the idea of using \textbf{STSM}s and how they help to improve the performance. Finally, we introduce a method to remove occluded points (\ref{occluded points culling by annnfs}) from the registration procedure.

\subsection{Tracking problem statement}
\label{sec:tracking problem statement}

\begin{figure}[t!]
  \centering
  \includegraphics[width=\linewidth]{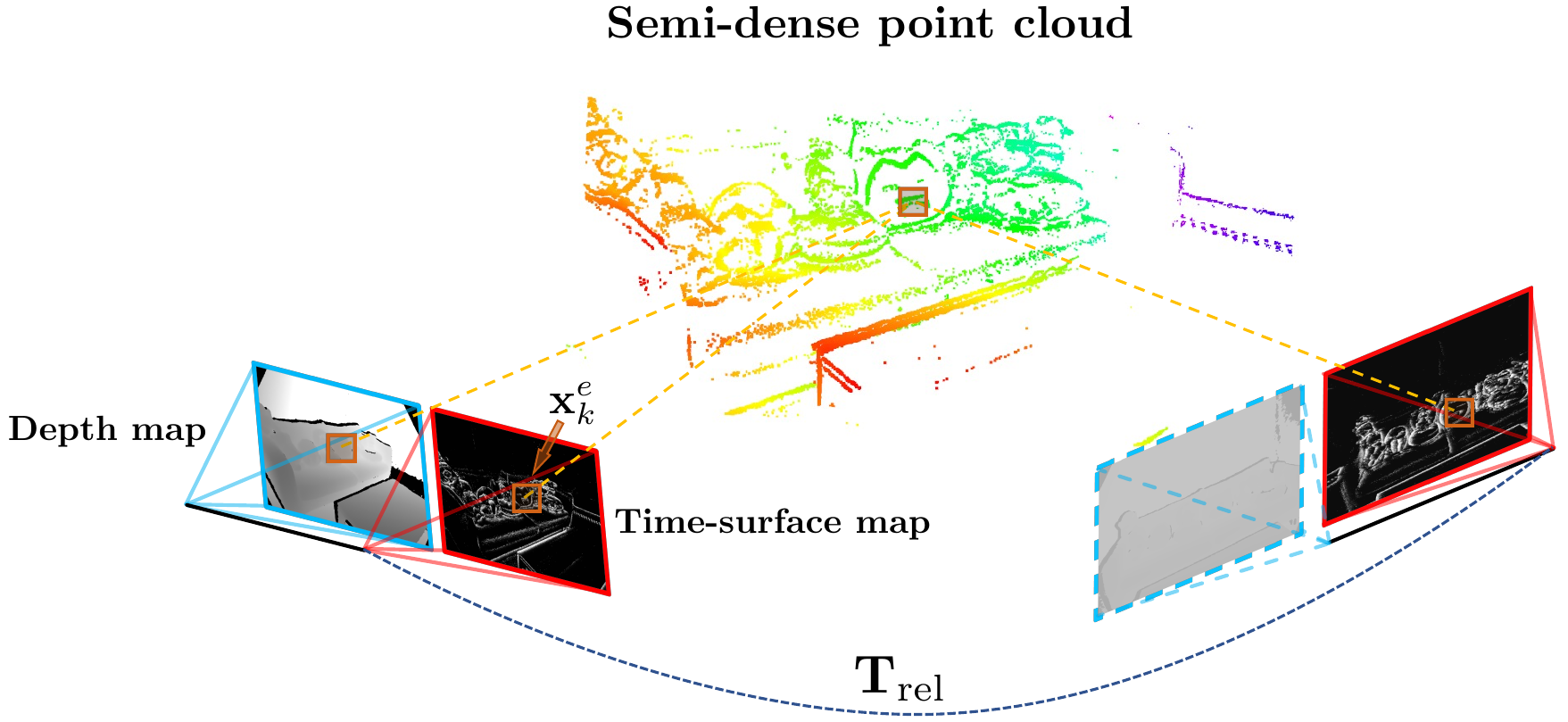}
  \caption{\textit{6-Dof Camera tracking}. Note that the depth image indicated in the dashed frame will not be used by the tracking module.}
  \label{fig:6dof_tracking}
  \vspace{-0.3cm}
\end{figure}

With the local or global 3D semi-dense point clouds from the mapping module in hand, we may now proceed to the details of our continuous, 6-DoF motion tracking module. We use the existing event-based localization strategy of Zhou et al. \cite{zhou2021event} in order to align subsequent \textbf{TSM}s with respect to the local semi-dense point cloud. As shown in Figure~\ref{fig:6dof_tracking}, \textbf{Canny-DEVO} runs a mapping thread to update the local map (i.e. the reference frame) each time the baseline with respect to the previous reference frame exceeds a given threshold. \textbf{Canny-EVT} in turn makes use of the global semi-dense map. In theory, given that events are triggered asynchronously and at very high rate (i.e. with temporal resolution in the order of micro-seconds), the pose of the camera can be updated at very high frequency.

The detailed form of the objective to be minimized is as follows. Let the local semi-dense 3D point cloud at a reference time $t_\text{ref}$ be given by $\mathcal{P}_{\text{ref}}$. The absolute pose of the current view is given by
\begin{equation}
\mathbf{T}(\boldsymbol{\theta}_\text{cur}) =\left[\begin{matrix}\mathbf{R}(\mathbf{q_\text{cur}}) & \mathbf{t}_\text{cur} \\ \mathbf{0}^\intercal & 1 \end{matrix}\right],
\end{equation}

\noindent where $\boldsymbol{\theta} = [\mathbf{t}^T \text{} \mathbf{q}^T]^T$ represents a motion parameter vector, $\mathbf{t}$ the position of the camera expressed in a world frame, and $\mathbf{q}$ its orientation as a Rodriguez vector. \textbf{Canny-DEVO} builds local maps online by using events and depth readings in nearby reference frames. In practice, we directly optimize the relative transformation from the current camera's position to a nearby reference frame defined by

\begin{small}
\begin{eqnarray}\small
\mathbf{T}_\text{rel}(\boldsymbol{\theta}_\text{rel}) & =  \mathbf{T}_\text{ref}(\boldsymbol{\theta}_\text{ref})^{-1}\mathbf{T}_\text{cur}(\boldsymbol{\theta}_\text{cur}).\nonumber
\end{eqnarray}
\end{small}
\noindent The optimization makes use of the function $W$ that transforms a 3D point $\mathbf{p}_{\text{ref}}\in\mathcal{P}_{\text{ref}}$ from the local map to the current frame. It is given by
\begin{equation}
W(\mathbf{p}_{\text{ref}};\boldsymbol{\theta}_\text{rel}) = \mathbf{T}^{-1}(\boldsymbol{\theta}_\text{rel})\cdot\mathbf{p}_{\text{ref}}.
\end{equation}
\textbf{Canny-EVT} in turn tracks a global semi-dense map, and we can therefore directly project points from the set $\mathcal{P}_\text{glo}$ of all 3D world points. In practice, we still find a nearby reference frame from the map and optimize the relative position w.r.t. to that pose. The corresponding tracking function is therefore given by
\begin{equation}
W(\mathbf{p}_{\text{glo}};\boldsymbol{\theta}_\text{rel}) = \mathbf{T}^{-1}(\boldsymbol{\theta}_\text{rel})\cdot\mathbf{T}^{-1}(\boldsymbol{\theta}_\text{ref})\cdot \mathbf{p}_{\text{glo}},
\end{equation}
where $\mathbf{p}_{\text{glo}}\in\mathcal{P}_{\text{glo}}$ is a global map point visible in the referenced, nearby keyframe.

The final goal of the tracking module is to find the optimal relative motion parameters $\boldsymbol{\theta}_\text{rel}$ that maximize the alignment of the reprojection of the local map $\mathcal{P}_{\text{ref}}$ or global map $\mathcal{P}_\text{glo}$ and the local minima in our current negated \textbf{TSM} \footnote{Note that the negated \textbf{TSM} does not refer to the \textbf{TSM} of the negative events, but the fact that the actual values of the \textbf{TSM} are sign-inverted to make sure that the values along the edges form local minimma rather than local maxima.}$\overline{\mathcal{T}}_\text{cur}(\cdot)$. Denoting both $\mathcal{P}_{\text{ref}}$ and $\mathcal{P}_\text{glo}$ as $\mathcal{P}$ and the warp functions as $W(\mathbf{p};\boldsymbol{\theta}_\text{rel})$, the objective function to find the optimum $\boldsymbol{\theta}_\text{rel}$ can be expressed as
\begin{equation}
\mathop{\arg\min}_{\boldsymbol{\theta}_\text{rel}}\sum_{\mathbf{p}_k\in \mathcal{P}}\rho(\overline{\mathcal{T}}_\text{cur}(\pi_e(W(\mathbf{p}_k;\boldsymbol{\theta}_\text{rel})))^2), \label{eq:objectFunc}
\end{equation}
where $\rho$ is a robust loss function. Similar to \cite{zhou2021event}, (\ref{eq:objectFunc}) is reformulated by using a forward compositional Lucas-Kanade method \cite{baker2004lucas}, which refines the incremental motion parameters $\Delta\boldsymbol{\theta}_\text{rel}$ by minimizing:
\begin{equation}
\mathop{\arg\min}_{\boldsymbol{\Delta\theta}_\text{rel}}\sum_{\mathbf{p}_k\in \mathcal{P}}
\rho(\overline{\mathcal{T}}_\text{cur}(\pi_e(W(W(\mathbf{p}_k;\Delta\boldsymbol{\theta}_\text{rel});\boldsymbol{\theta}_\text{rel})))^2),
\label{eq:objectFunc2}
\end{equation}
and the new warping function $W(W(\mathbf{p}_k;\Delta\boldsymbol{\theta}_\text{rel});\boldsymbol{\theta}_\text{rel})$ is updated in each iteration. The new compositional approach is more efficient than the original method given that the Jacobian of the objective function remains constant at the position of zero increment and can be pre-computed. Smoothness, differentiability and convexity of this method are proven in \cite{zhou2021event}. 

\subsection{Exploting \textbf{TSM}s as distance fields}
\label{exploting tsm as df}

The tracking proceeds by constructing a potential field in the current view. The \textbf{TSM} encodes the motion history of the appearance edges. The current edge locations have higher values than the previous edge locations. Conversely, the negated \textbf{TSM} shows smaller values along the edges of the current frame. The potential field is constructed by negating and offsetting the \textbf{TSM} at the current time $t_\text{cur}$, i.e. $\overline{\mathcal{T}}_\text{cur}(\cdot) = 1 -\mathcal{T}(\cdot,t_\text{cur})$. \cite{zhou2021event} demonstrates how this field can in fact be interpreted as a distance field, and we may readily use existing, distance field-based geometric semi-dense registration techniques. Based on a hypothesized pose, the projected point locations from the semi-dense point cloud lead to a sampling of the field, and the sum of squares of the sampled values is considered as an energy to be minimized over the pose parameters of the camera.

\subsection{Polarity-aware registration by \textbf{STSMs}}
\label{polarity-aware registration by STSMs}

\begin{figure}[t!]
\centering
\subfigure[Simple quantization]{\label{predict polarity1}\includegraphics[width=0.4\linewidth]{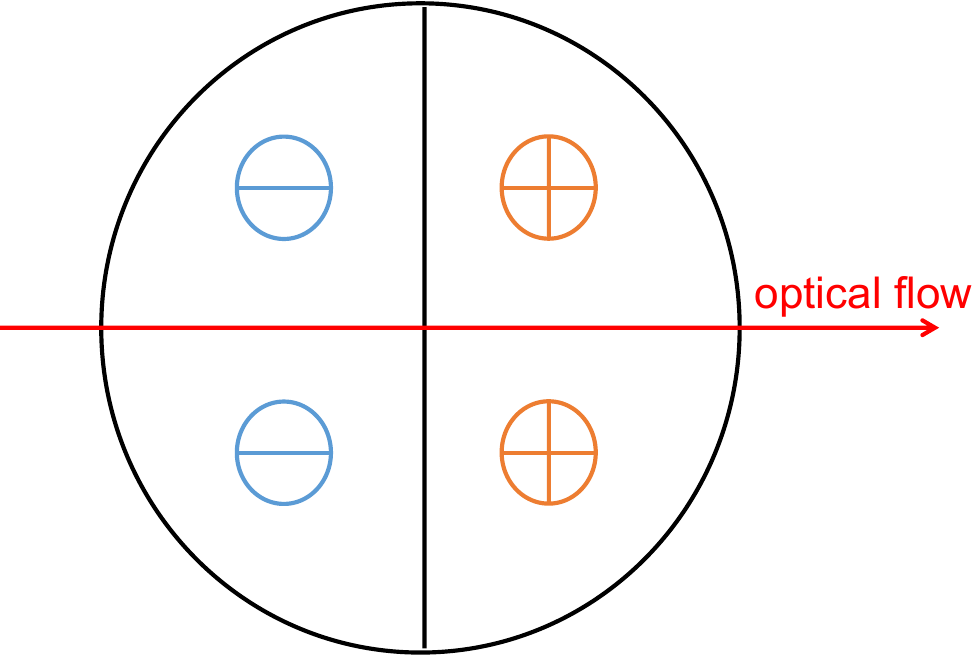}}
\centering
\subfigure[Quantization with neutral buffers]{\label{predict polarity2}\includegraphics[width=0.4\linewidth]{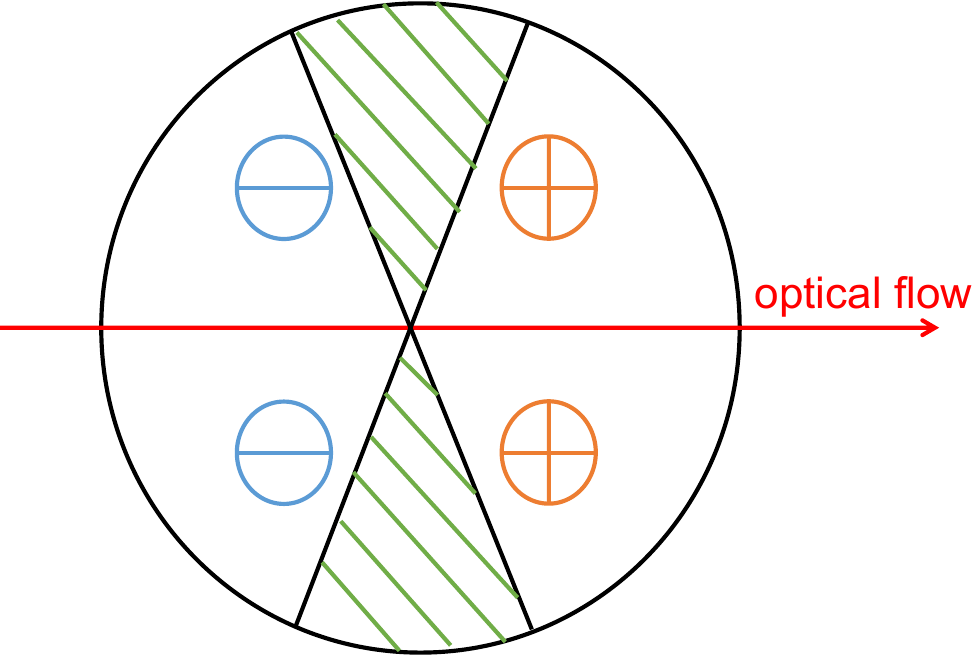}}
\caption{\label{predict polarity} \textit{Visualization of the quantization used for predicting the polarity of events.}}
\end{figure}

\begin{figure}[b!]
\centering
\subfigure[Normal registration]{\label{energy4}\includegraphics[width = 0.158\textwidth]{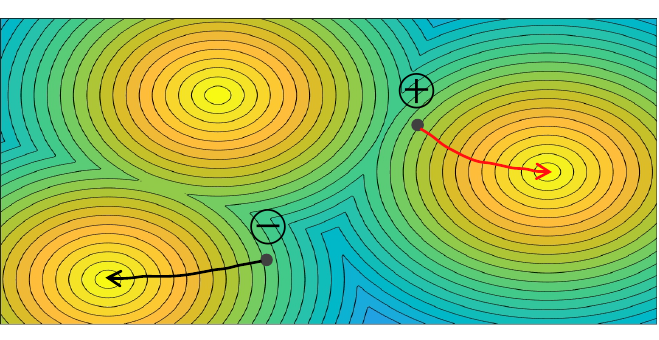}}
\centering
\subfigure[Polarity-aware registration for positive point]{\label{energy5}\includegraphics[width = 0.158\textwidth]{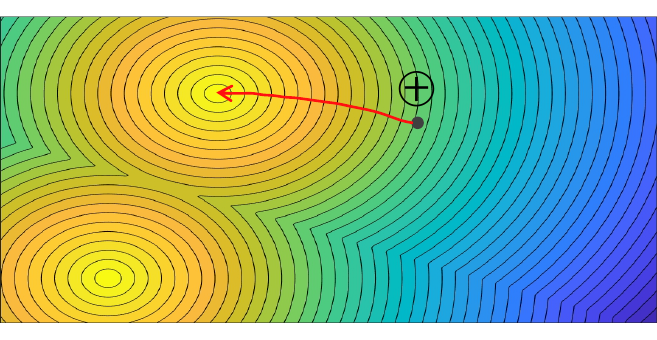}}
\centering
\subfigure[Polarity-aware registration for negative point]{\label{energy6}\includegraphics[width = 0.158\textwidth]{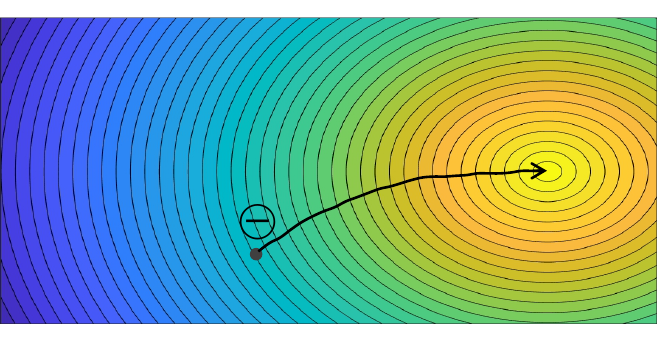}}
\caption{\label{energy}\textit{Visualization of the potential field given by a small set of event measurements}: \textit{Left}: \textbf{TSM} with two local minima on the left generated by positive events, and one local minimum on the right generated by a negative event. As can be observed, large displacements may easily cause the reprojected points to fall into the basin of attraction of the wrong local minimum, thus causing overall tracking failure. \textit{Center and Right}: If we use polarity-aware \textbf{STSM}s, the reprojected points with their predicted polarities are easily assigned to the correct basins of attraction.}
\end{figure}

\textbf{Canny-VO}~\cite{zhou2018canny} orients and splits the distance fields according to a quantization of gradient vectors to avoid registration biases due to partial observations in the model-to-data registration approach. The method furthermore helps in enlarging the convergence basin during registration, as for example registration to the wrong-side edge in the presence of thin lines (i.e. lines with two adjacent edges with opposite gradient directions) is simply avoided. Inspired by \textbf{Canny-VO}~\cite{zhou2018canny}, we follow a similar approach and split our \textbf{TSM} based on polarity, thus resulting in Signed Time Surface Maps (\textbf{STSM}s). During registration, we then predict the polarity of an event triggered by a reprojected point by projecting the 3D gradient into the new view and also estimating optical flow by combining the point depth and a propagated camera velocity (e.g. using the constant velocity motion model). The inner product of both entities indicates the normal flow. The red vector in Figure~\ref{predict polarity} is the direction of the predicted optical flow. We predict a positive event if the inscribed angle between the reprojected 3D gradient vector and the optical flow vector is smaller than a certain degree (e.g. 90$^{\circ}$ in Figure~\ref{predict polarity1}), and otherwise predict a negative event.

In practice, the predicted optical flow and reprojected gradient vectors cannot be obtained without any errors. In order to make the system more robust, we insert a buffer area between the positive and negative areas (shaded in green in Figure~\ref{predict polarity2}). If the reprojected gradient vector falls into the buffer area, we consider the predicted polarity of the point to be not trust-worthy, and continue to use the regular \textbf{TSM} in which all events have been considered, irrespectively of their polarity. For other points, we consider the predicted polarity to be trust-worthy, and use \textbf{STSM}s. As illustrated in Figure~\ref{energy}, the distance to the corresponding regions of attraction of specific events can be much larger in the case of \textbf{STSM}s than in the case of \textbf{TSM}s. \revisetwo{The buffer area we set is 30$^{\circ}$. It is helpful to counteract the impact of the wrong prediction of the polarity. The effect of the size of the buffer area on the accuarcy is not sensitive. Simply removing the events in the buffer area has no substantial performance impact.} An example on real data is shown in Fig.~\ref{STSM}.

\subsection{Occlusion handling using \textbf{ANNFs}}
\label{occluded points culling by annnfs}

\begin{figure}[t!]
\centering
\subfigure[Distance Field]{\label{distance fields}\includegraphics[height = 0.169\textwidth]{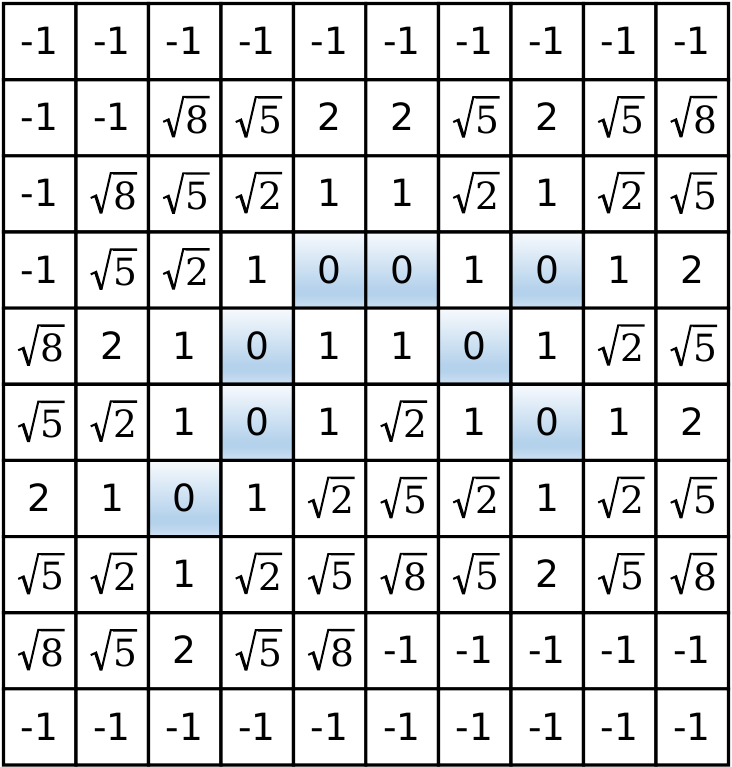}}
\centering
\subfigure[Approximate Nearest Neighbor Field]{\label{annf}\includegraphics[height = 0.18\textwidth]{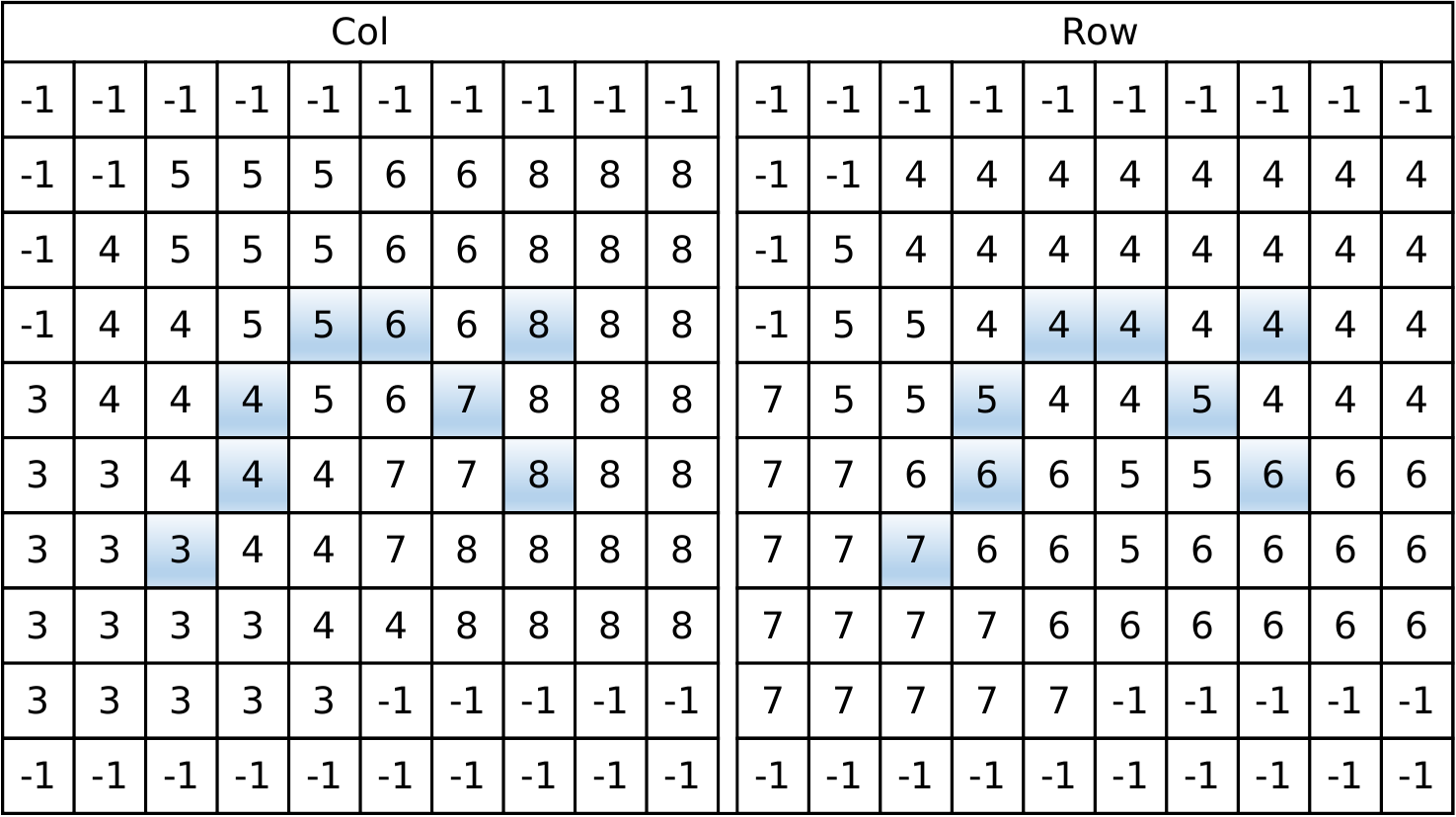}}
\caption{\label{annfs} \textit{Example of a distance field and the corresponding approximate nearest neighbor field}. Blue fields indicate pixels where an edge point has been detected. Distance fields indicate the distance to the nearest point. The approximate nearest neighbor field instead indicates the row and column of the nearest point on the edge.}
\end{figure}

\begin{figure}[b!]
  \centering
  \includegraphics[width=0.4\linewidth]{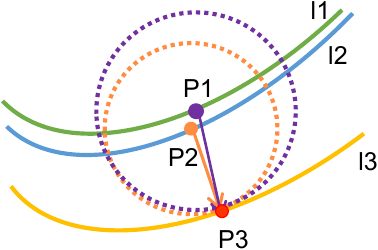}
  \caption{\textit{Illustration of the principle registration conflict handling mechanism using \textbf{ANNFs}}: The green curve (l1) corresponds to the reprojection of occluded points, whereas the blue curve (l2) corresponds to the reprojection of visible points. The yellow curve (l3) corresponds to the location of these points in the current frame. Both the purple point (P1) from l1 and the orange point (P2) from l2 share the same nearest neighbor point P3 from the observed edge. In case of such a conflict, we will compare the depths of P1 and P2. Only the point with smaller depth is added to the registration.}
  \label{fig:point culling}
\end{figure}

\textbf{Canny-VO} introduces approximate nearest neighbor fields to remove problems related to discontinuities or nonsmoothness during energy minimization. As explained in Figure~\ref{annf}, ANNFs communicate the position of the nearest point along an edge. In this paper, we exploit a new application of \textbf{ANNFs} for occlusion handling.

Our event-based tracking relies on registering 3D points in our \textbf{TSM}. There is two kinds of points that should potentially be ignored. The first kind is given by occluded points. When building the semi-dense map, the system potentially observed the scene from different directions. Given the semi-dense nature of the map, it is easily possible that occluded points will project into the current frame (see Figure~\ref{repj_occluded} for an example on real data). The second kind is given by points that cannot trigger any events as the reprojected gradient is perpendicular to the velocity direction. In this case, the point in question will fail to trigger any events, and thus its reprojection may land in an area of the potential field that has no corresponding local minmimum. We add a simple heuristic to reduce the consideration of such points in our energy function. We use an ANNF (cf. Figure~\ref{annfs}) in order to register reprojected points to their corresponding edge point. This will permit the detection and avoidance of collisions given by multiple 3D points registering to the same edge point in the current frame. If such collisions happen, we will notably limit the registered points to the closest one. The principle of this point culling mechanism is explained in Figure~\ref{fig:point culling}.

\section{Experimental Evaluation}\label{sec:experiments}

We evaluate our novel visual odometry and tracking frameworks on both public and self-collected sequences. We start by introducing our experiment implementation and the used hardware configuration. Next, we compare \textbf{Canny-DEVO} against several alternatives on both mild test sequences and more challenging scenarios. All compared alternatives are odometry-style frameworks that do not perform loop closure. The compared alternatives are given by state-of-the-art event-based and RGB-D or depth-only approaches, including the regular vision alternative \textbf{Canny-VO}. Next, we compare \textbf{Canny-VT}---a regular vision-based tracker of the global semi-dense map---against \textbf{Canny-EVT} to demonstrate stable and accurate cross-modal tracking across a large variety of challenging scenarios using only a single event camera. We further investigate the advantages of \textbf{STSM}s over \textbf{TSM}s by comparing the performance for different camera framerates and illumination conditions. Both qualitative and quantitative results are provided, which demonstrate the effectiveness of our method. We also provide an analysis of the computational performance of all systems\revise{, as well as a concluding discussion in which the importance of the individual additions to our tracking module, as well as the cross-modal approach itself are further analyzed.} The content of our published open-source software framework is summarized in Table~\ref{tab:canny-voat}.


\begin{table}[b]
  \centering
  \caption{Overview of the methods supported by our open-source released framework (\href{https://github.com/zyfff/Canny-EVT/}{https://github.com/zyfff/Canny-EVT/}).}
    \begin{tabular}{c|cc}
    \toprule
          & \multicolumn{2}{c}{\textbf{Canny-VOAT}} \\
    \midrule
    \multicolumn{1}{c|}{\multirow{2}[2]{*}{\diagbox[dir=NW]{Sensors}{Methods}}} & Visual odometry & Visual tracking \\
          & (online local mapping) & (offline global mapping) \\
    \midrule
    Normal camera & \textbf{Canny-VO} & \textbf{Canny-VT} \\
    Event camera & \textbf{Canny-DEVO} & \textbf{Canny-EVT} \\
    \bottomrule
    \end{tabular}%
  \label{tab:canny-voat}%
\end{table}%

\subsection{Datasets and Sensor Setup}
\label{experimental setup and dataset used}

\begin{figure}[t!]
	\centering
	\includegraphics[width=0.70\linewidth]{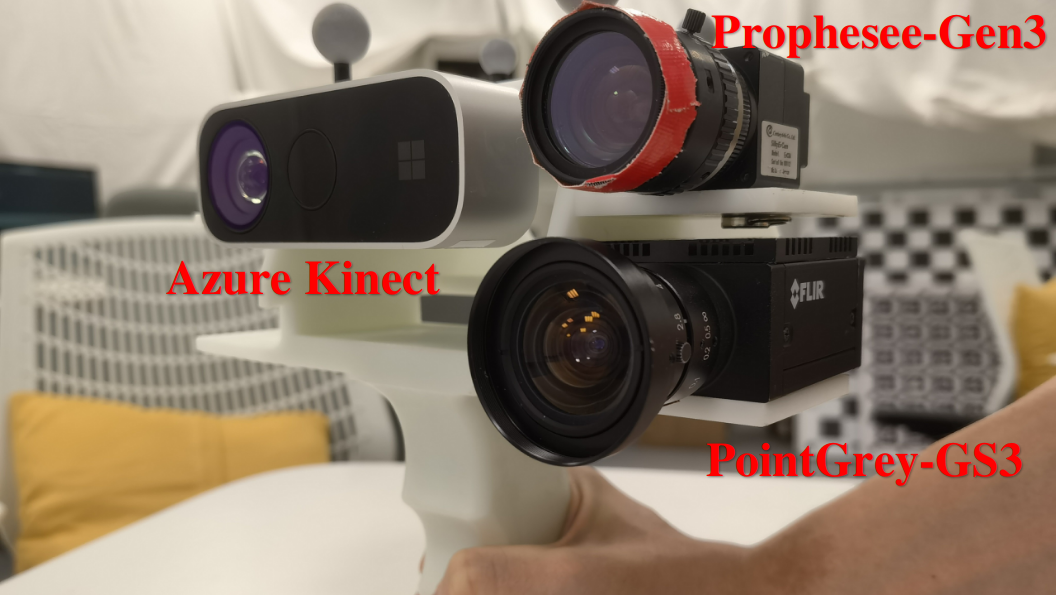}
	\caption{Custom sensor system with event camera, regular camera, and RGB-D sensor for the self-collected datasets.}
	\label{fig:hardware_setup}
\end{figure}

Our first experiments are conducted on the Multi-Vehicle-Stereo-Event-Camera dataset (\textit{MVSEC}) presented in \cite{2018The}. These publicly available sequences include synchronized event streams, intensity images and depth images with ground truth trajectories. We add sequences from further public datasets, including \textit{VECtor}~\cite{gao2022vector}\revise{ and TUM~\cite{sturm2012benchmark}}. Finally, in order to put a full stress test onto all methods, we test the methods on several other, self-collected sequences with different types of textures, motion characteristics, and illumination conditions. For different types of scene textures, the sequences are named \textit{cali}, \textit{table} and \textit{sofa}, respectively. \textit{cali} is a scene with many calibration boards, \textit{table} a standard desktop environment, and \textit{sofa} a living room scene. For each texture, we capture datasets under three different motion speeds, denoted \textit{fast}, \textit{mid} and \textit{slow}. More datasets are captured under a variety of illumination conditions, denoted \textit{bright}, \textit{darkish}, \textit{dim}, \textit{dark}, and \textit{hdr}. All sequences are listed in Table \ref{tab:all comparison}. The sequences are collected by a custom-designed, hardware-synchronized multi-sensor system (cf. Figure~\ref{fig:hardware_setup}), which contains a global-shutter industrial camera (PointGrey-GS3), a high-resolution event camera (Prophesee-Gen3), and an RGB-D sensor (Azure Kinect). Detailed specifications are listed in Table \ref{tab:specifications}. The multi-sensor system is intrinsically and extrinsically calibrated, and ground truth for all sequences is captured by a highly accurate external motion capture system.

\begin{table}[t!]
  \captionsetup{justification=centering}
  \centering
  \caption{Specifications of sensors used in the custom sensor setup.}
  \label{tab:Setup}
  \setlength{\tabcolsep}{5.8pt}
    \begin{tabular}{cc|cccccc}
    \toprule
    \multicolumn{2}{c|}{Sensor} & \multicolumn{2}{c}{Exposure Time} & \multicolumn{2}{c}{Resolution} & \multicolumn{2}{c}{Frame Rate} \\
    \midrule
    \multicolumn{2}{c|}{PointGrey-GS3} & \multicolumn{2}{c}{10ms} & \multicolumn{2}{c}{1224$\times$1024} & \multicolumn{2}{c}{30fps} \\
    \multicolumn{2}{c|}{Azure Kinect} & \multicolumn{2}{c}{12.8ms} & \multicolumn{2}{c}{640$\times$576} & \multicolumn{2}{c}{30fps} \\
    \multicolumn{2}{c|}{Prophesee-Gen3} & \multicolumn{2}{c}{-} & \multicolumn{2}{c}{640$\times$480} & \multicolumn{2}{c}{-} \\
    \bottomrule
    \end{tabular}%
  \label{tab:specifications}%
\end{table}

\subsection{Comparison of \textbf{Canny-DEVO} against alternatives}
\label{compare_Canny-DEVO_with_RGBD}


\subsubsection{Comparison against purely event-based solutions}
\label{compare_with_RGBD}

We first compare our proposed depth-event method \textbf{Canny-DEVO}~\cite{zuo2022devo} against \textbf{ESVO}, an open-source event-based stereo visual odometry framework published in \cite{zhou2021event}. The two methods are evaluated on the public dataset \textit{MVSEC}~\cite{2018The}. We choose both indoor and outdoor sequences, which are captured by a flying drone inside a room, and a stereo event camera mounted on a vehicle, respectively. Note that the depth measurements in \textit{MVSEC} are obtained from a LiDAR, which can easily be considered as a replacement for the depth camera in our method.


\begin{table}[t]
\captionsetup{justification=centering}
\centering
  \caption{Comparison of \textbf{Canny-DEVO} against a pure event-based method on \textit{MVSEC}\\ $\left[\mathbf{R}_\text{\upshape rpe}\text{\upshape : °/s}, \mathbf{t}_\text{\upshape rpe}\text{\upshape : cm/s}, \mathbf{t}_\text{\upshape ate}\text{\upshape : cm}\right]$}
  \label{tab:ESVO Comparison}
  \renewcommand\arraystretch{1.3}
 \setlength{\tabcolsep}{3.7pt}
\begin{tabular}{llccccccc}
\toprule
                              &  & \multicolumn{3}{c}{\textbf{Canny-DEVO}}              &  & \multicolumn{3}{c}{\textbf{ESVO}} \\ \cline{3-5} \cline{7-9} 
\textit{Sequence}             &  & $\mathbf{R}_\text{rpe}$          & $\mathbf{t}_\text{rpe}$          & $\mathbf{t}_\text{ate}$           &  & $\mathbf{R}_\text{rpe}$       & $\mathbf{t}_\text{rpe}$       & $\mathbf{t}_\text{ate}$      \\ \cline{1-5} \cline{7-9} 
\textit{upenn indoor flying1} &  & \textbf{0.30} & \textbf{0.88} & \textbf{20.58} &  & 0.37      & 1.63      & 21.68     \\
\textit{upenn indoor flying2} &  & \textbf{0.36} & \textbf{1.12} & \textbf{11.33} &  & -         & -         & -         \\
\textit{upenn indoor flying3} &  & \textbf{0.53} & \textbf{1.21} & \textbf{10.60} &  & 0.54      & 2.14      & 25.40     \\
\textit{upenn indoor flying4} &  & \textbf{0.53} & \textbf{1.44} & \textbf{13.16} &  & -         & -         & -         \\
\textit{upenn outdoor day1}   &  & \textbf{0.30} & \textbf{7.77} & \textbf{88.70} &  & -         & -         & -         \\ \bottomrule
\end{tabular}
\end{table}

\begin{figure}[b]
\centering
\subfigure[Scene]{\includegraphics[width=4.0cm]{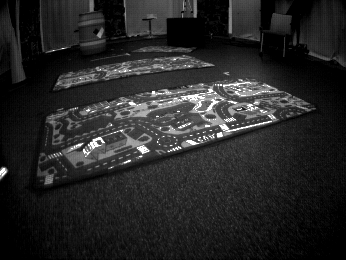}}
\subfigure[Time Surface Map]{\includegraphics[width=4.0cm]{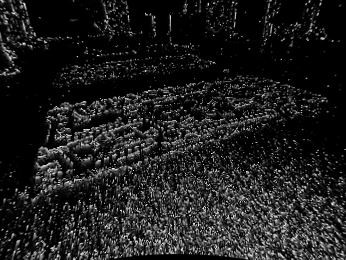}}
\\ 
\centering
\subfigure[Reprojection Map by~\cite{zhou2021event}]{\includegraphics[width=4.0cm]{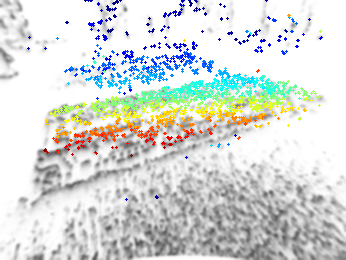}}
\subfigure[Reprojection Map (Ours)]{\includegraphics[width=4.0cm]{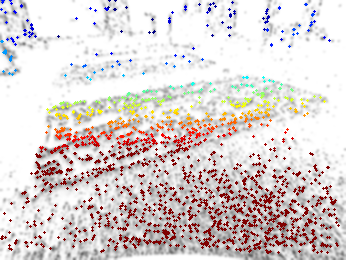}}
\caption{\label{MVSEC_figure_comparison}Visualization of an image (top left) and \textbf{TSM} (top right) from sequence \textit{indoor flying2} and corresponding reprojections into a nearby frame for \textbf{ESVO} (bottom left) and \textbf{Canny-DEVO} (bottom right). The coloring indicates the depth of each point.}
\end{figure}

Quantitative results are listed in Table \ref{tab:ESVO Comparison}. As can be observed, \textbf{Canny-DEVO} clearly outperforms \textbf{ESVO} in all sequences. It should be noted that sequences \textit{indoor flying2} and \textit{indoor flying4} are much more challenging than the other two sequences owing to high noise in the event streams caused by a combination of difficult texture and highly dynamic platform motion. Examples of the sequence are indicated in Figure~\ref{MVSEC_figure_comparison}. The ground surface triggers a large number of noisy events for which depth is hard to observe. This severely influences the mapping result of \textbf{ESVO} and---due to the highly interleaved tracking and mapping modules---causes tracking failures in this fast exploration scenario. In contrast, \textbf{Canny-DEVO} directly reads the depth from the depth sensor, the quality of which is less influenced by the noisy nature of the texture. The independent depth readings significantly contribute to the robustness of the entire system when the inputs of the event camera degrade. Furthermore, the ability of any stereo method to perceive depths beyond a certain range is limited by the baseline of the system, which is why \textbf{ESVO} is unable to provide competitive results on the outdoor sequence.

\begin{table}[h]
  \centering
  \caption{Capabilities for different methods under different lighting conditions}
  \label{tab:MPL Availability}
  \setlength{\tabcolsep}{7pt}
    \begin{tabular}{cc|cccccccc}
    \toprule
    \multicolumn{2}{c|}{Sequence} & \multicolumn{2}{c}{\textbf{Canny-DEVO}} & \multicolumn{2}{c}{\textbf{KinectFusion}~\cite{newcombe2011kinectfusion}} & \multicolumn{2}{c}{\textbf{Canny-VO}~\cite{zhou2018canny}}  \\
    \midrule
    \multicolumn{2}{c|}{\textit{light}} & \multicolumn{2}{c}{\textcolor[rgb]{ 0,  .69,  .314}{\Checkmark}} & \multicolumn{2}{c}{\textcolor[rgb]{ 0,  .69,  .314}{\Checkmark}} & \multicolumn{2}{c}{\textcolor[rgb]{ 0,  .69,  .314}{\Checkmark}} \\
    \multicolumn{2}{c|}{\textit{darkish}} & \multicolumn{2}{c}{\textcolor[rgb]{ 0,  .69,  .314}{\Checkmark}} & \multicolumn{2}{c}{\textcolor[rgb]{ 0,  .69,  .314}{\Checkmark}} & \multicolumn{2}{c}{\textcolor[rgb]{ 1,  0,  0}{\XSolidBrush}}  \\
    \multicolumn{2}{c|}{\textit{dim}} & \multicolumn{2}{c}{\textcolor[rgb]{ 0,  .69,  .314}{\Checkmark}} & \multicolumn{2}{c}{\textcolor[rgb]{ 0,  .69,  .314}{\Checkmark}} & \multicolumn{2}{c}{\textcolor[rgb]{ 1,  0,  0}{\XSolidBrush}}  \\
    \multicolumn{2}{c|}{\textit{dark}} & \multicolumn{2}{c}{\textcolor[rgb]{ 0,  .69,  .314}{\Checkmark}} & \multicolumn{2}{c}{\textcolor[rgb]{ 0,  .69,  .314}{\Checkmark}} & \multicolumn{2}{c}{\textcolor[rgb]{ 1,  0,  0}{\XSolidBrush}} \\
    \multicolumn{2}{c|}{\textit{HDR}} & \multicolumn{2}{c}{\textcolor[rgb]{ 0,  .69,  .314}{\Checkmark}} & \multicolumn{2}{c}{\textcolor[rgb]{ 0,  .69,  .314}{\Checkmark}} & \multicolumn{2}{c}{\textcolor[rgb]{ 1,  0,  0}{\XSolidBrush}} \\
    \bottomrule
    \end{tabular}%
  \label{tab:addlabel}%
\end{table}%

\begin{table}[t]
\captionsetup{justification=centering}
  \centering
  \caption{Comparison for different depth frame rates\\ $\left[\mathbf{R}_\text{\upshape rpe}\text{\upshape : °/s}, \mathbf{t}_\text{\upshape rpe}\text{\upshape : cm/s}, \mathbf{t}_\text{\upshape ate}\text{\upshape : cm}\right]$}
  \label{tab:Frequency}
  \setlength{\tabcolsep}{1.9pt}
    \renewcommand\arraystretch{1.25}
\begin{tabular}{ccccccccccccc}
\toprule
\textit{Frequency} & \multicolumn{1}{l}{} & \multicolumn{3}{c}{\textbf{Canny-DEVO}}                & \multicolumn{1}{l}{\textbf{}} & \multicolumn{3}{c}{\textbf{Canny-VO}} & \multicolumn{1}{l}{\textbf{}} & \multicolumn{3}{c}{\textbf{KinectFusion}} \\ \cline{3-5} \cline{7-9} \cline{11-13} 
\textit{Fast}      &                      & $\mathbf{R}_\text{rpe}$           & $\mathbf{t}_\text{rpe}$            &  $\mathbf{t}_\text{ate}$          &                               & $\mathbf{R}_\text{rpe}$ & $\mathbf{t}_\text{rpe}$          & $\mathbf{t}_\text{ate}$            &                               & $\mathbf{R}_\text{rpe}$           & $\mathbf{t}_\text{rpe}$    & $\mathbf{t}_\text{ate}$            \\ \cline{1-5} \cline{7-9} \cline{11-13} 
30                 &                      & 1.50           & 2.42           & 46.37          &                               & 1.51 & \textbf{2.22} & 30.81          &                               & \textbf{1.37}   & 2.75   & \textbf{27.00}  \\
15                 &                      & \textbf{2.92}  & 4.86           & 46.70          &                               & 2.96 & \textbf{4.46} & \textbf{29.11} &                               & 3.04            & 8.36   & 37.99           \\
10                 &                      & \textbf{4.26}  & 7.32           & 47.92          &                               & 4.65 & \textbf{6.55} & \textbf{34.68} &                               & 4.83            & 17.33  & 66.78           \\
5                  &                      & \textbf{7.73}  & \textbf{14.73} & 57.89          &                               & -    & -             & -              &                               & 9.01            & 26.86  & \textbf{55.09}  \\
1                  &                      & \textbf{18.58} & \textbf{49.16} & \textbf{76.04} &                               & -    & -             & -              &                               & -               & -      & -               \\ \cline{1-5} \cline{7-9} \cline{11-13} 
\textit{Medium}    &                      & $\mathbf{R}_\text{rpe}$          & $\mathbf{t}_\text{rpe}$            & $\mathbf{t}_\text{ate}$          &                               & $\mathbf{R}_\text{rpe}$  & $\mathbf{t}_\text{rpe}$           & $\mathbf{t}_\text{ate}$            &                               & $\mathbf{R}_\text{rpe}$             & $\mathbf{t}_\text{rpe}$   & $\mathbf{t}_\text{ate}$             \\ \cline{1-5} \cline{7-9} \cline{11-13} 
30                 &                      & 1.16           & 1.53           & 27.83          &                               & 1.18 & \textbf{1.25} & \textbf{19.68} &                               & \textbf{1.10}   & 1.79   & 21.71           \\
15                 &                      & 2.29           & 3.01           & 24.20          &                               & 2.34 & \textbf{2.48} & \textbf{20.08} &                               & \textbf{2.17}   & 3.56   & 21.31           \\
10                 &                      & 3.39           & 4.51           & 21.46          &                               & 3.49 & \textbf{3.70} & \textbf{20.34} &                               & \textbf{3.20}   & 5.58   & 58.73           \\
5                  &                      & \textbf{6.55}  & \textbf{9.20}  & \textbf{21.55} &                               & -    & -             & -              &                               & 7.16            & 15.80  & 37.07           \\
1                  &                      & \textbf{18.46} & \textbf{35.24} & \textbf{51.93} &                               & -    & -             & -              &                               & -               & -      & -               \\ \cline{1-5} \cline{7-9} \cline{11-13} 
\textit{Slow}      &                      & $\mathbf{R}_\text{rpe}$         &$\mathbf{t}_\text{rpe}$          & $\mathbf{t}_\text{ate}$           &                               & $\mathbf{R}_\text{rpe}$ & $\mathbf{t}_\text{rpe}$          & $\mathbf{t}_\text{ate}$           &                               & $\mathbf{R}_\text{rpe}$            & $\mathbf{t}_\text{rpe}$   & $\mathbf{t}_\text{ate}$            \\ \cline{1-5} \cline{7-9} \cline{11-13} 
30                 &                      & 0.63           & 1.00           & \textbf{19.50} &                               & 0.64 & \textbf{0.82} & 26.94          &                               & \textbf{0.59}   & 1.15   & 22.48           \\
15                 &                      & 1.21           & 1.97           & \textbf{18.44} &                               & 1.24 & \textbf{1.65} & 26.01          &                               & \textbf{1.13}   & 2.29   & 22.22           \\
10                 &                      & 1.76           & 2.98           & \textbf{18.46} &                               & 1.82 & \textbf{2.46} & 26.32          &                               & \textbf{1.66}   & 3.41   & 22.10           \\
5                  &                      & \textbf{3.28}           & 6.09           & \textbf{17.61} &                               & 3.71 & \textbf{5.54} & 27.97          &                               & -   & -  & -          \\
1                  &                      & \textbf{10.66} & \textbf{31.15} & \textbf{37.88} &                               & -    & -             & -              &                               & -               & -      & -               \\ \bottomrule
\end{tabular}
\vspace{-0.2cm}
\end{table}

\begin{table*}[t!]
\vspace{0.2cm}
\centering
\captionsetup{justification=centering}
\caption{RPE and ATE on self-collected datasets for \textbf{Canny-DEVO} and alternatives \\$\left[\mathbf{R}_\text{\upshape rpe}\text{\upshape : °/s}, \mathbf{t}_\text{\upshape rpe}\text{\upshape : cm/s}, \mathbf{t}_\text{\upshape ate}\text{\upshape : cm}\right]$}
  \label{tab:all comparison}
  \setlength{\tabcolsep}{9.7pt}
  \renewcommand\arraystretch{1.2}
\begin{tabular}{lcccccccccccc}
\toprule
                             & \multicolumn{1}{c}{} & \multicolumn{3}{c}{\textbf{Canny-DEVO}}                      & \multicolumn{1}{c}{} & \multicolumn{3}{c}{\textbf{Canny-VO}}                   &  & \multicolumn{3}{c}{\textbf{KinectFusion}}              \\ \cline{3-5} \cline{7-9} \cline{11-13} 
\textit{Sequence}            &                       & $\mathbf{R}_\text{rpe}$          & $\mathbf{t}_\text{rpe}$          & $\mathbf{t}_\text{ate}$           &                       & $\mathbf{R}_\text{rpe}$          & $\mathbf{t}_\text{rpe}$          & $\mathbf{t}_\text{ate}$           &  & $\mathbf{R}_\text{rpe}$          & $\mathbf{t}_\text{rpe}$          & $\mathbf{t}_\text{ate}$           \\ \cline{1-5} \cline{7-9} \cline{11-13} 
\textit{cali\_bright\_fast}  &                       & \textbf{3.73} & 2.03          & 23.67          &                       & 3.81          & \textbf{1.47} & 15.55          &  & 3.74          & 1.81          & \textbf{15.34} \\
\textit{cali\_bright\_mid}   &                       & 1.44          & 1.77          & \textbf{16.90} &                       & 1.42          & \textbf{1.26} & 21.23          &  & \textbf{1.35} & 1.77          & 19.22          \\
\textit{cali\_bright\_slow}  &                       & \textbf{0.97} & 0.78          & 11.85          &                       & 1.03          & \textbf{0.59} & \textbf{7.16}  &  & 0.99          & 0.89          & 14.52          \\
\textit{cali\_darkish\_slow} &                       & 1.03          & \textbf{0.91} & 18.02          &                       & -             & -             & -              &  & \textbf{1.02} & 0.93          & \textbf{11.34} \\
\textit{cali\_dim\_slow}     &                       & \textbf{1.55} & 0.88          & 35.38          &                       & -             & -             & -              &  & 1.62          & \textbf{0.82} & \textbf{9.05}  \\
\textit{cali\_dark\_fast}    &                       & \textbf{0.58} & \textbf{0.87} & 26.43          &                       & -             & -             & -              &  & 0.63          & 0.92          & \textbf{12.61} \\
\textit{cali\_dark\_mid}     &                       & \textbf{0.49} & 0.60          & 17.65          &                       & -             & -             & -              &  & 0.54          & \textbf{0.59} & \textbf{12.89} \\
\textit{cali\_dark\_slow}    &                       & \textbf{0.24} & 0.31          & 9.85           &                       & -             & -             & -              &  & 0.26          & \textbf{0.23} & \textbf{8.97}  \\
\textit{cali\_hdr\_slow}     &                       & \textbf{0.92} & 0.79          & 21.55          &                       & -             & -             & -              &  & 0.95          & \textbf{0.71} & \textbf{11.10} \\
\textit{table\_bright\_fast} &                       & 1.50          & 2.42          & 46.37          &                       & 1.51          & \textbf{2.22} & 30.81          &  & \textbf{1.38} & 2.75          & \textbf{27.00} \\
\textit{table\_bright\_mid}  &                       & 1.16          & 1.53          & 27.83          &                       & 1.18          & \textbf{1.25} & \textbf{19.68} &  & \textbf{1.10} & 1.79          & 21.71          \\
\textit{table\_bright\_slow} &                       & 0.63          & 1.00          & \textbf{19.5}  &                       & 0.64          & \textbf{0.82} & 26.94          &  & \textbf{0.59} & 1.15          & 22.48          \\
\textit{sofa\_bright\_fast}  &                       & \textbf{2.60} & 2.30          & 30.22          &                       & 2.63          & \textbf{1.90} & \textbf{23.89} &  & 2.61          & 3.64          & 27.79          \\
\textit{sofa\_bright\_mid}   &                       & 5.28          & 4.02          & \textbf{13.4}  &                       & \textbf{1.18} & \textbf{1.25} & 19.68          &  & 3.13          & 7.62          & 71.5           \\
\textit{sofa\_bright\_slow}  &                       & 1.47          & 1.16          & \textbf{10.94} &                       & \textbf{0.64} & \textbf{0.82} & 26.94          &  & 1.47          & 1.21          & 21.82          \\ \bottomrule
\end{tabular}
\end{table*}
\begin{figure*}[b]
\vspace{-0.2cm}
\centering
\subfigure[Scene]{\includegraphics[height = 0.21\textwidth]{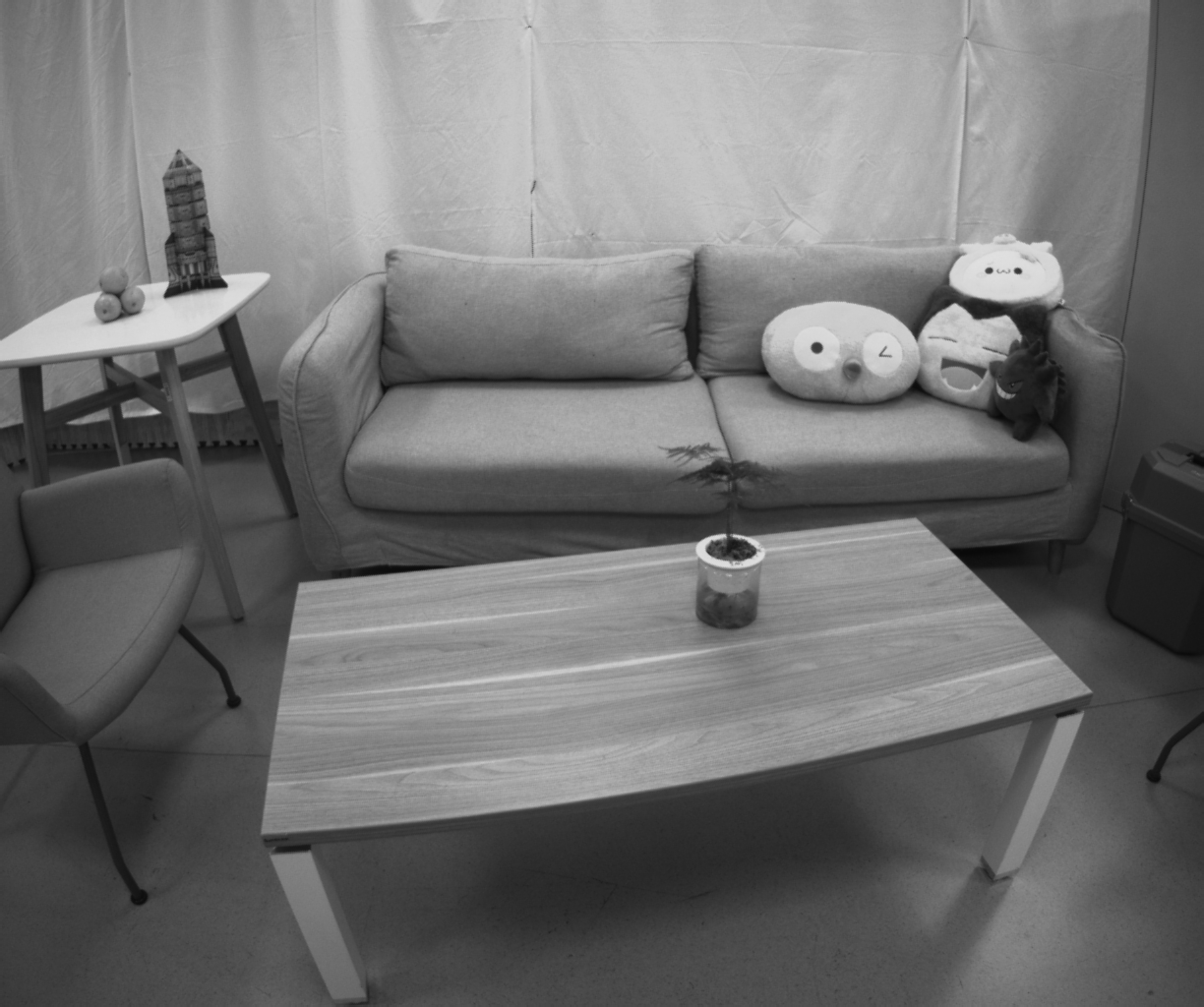}}
\subfigure[Global Semi-dense Map]{\includegraphics[height = 0.21\textwidth]{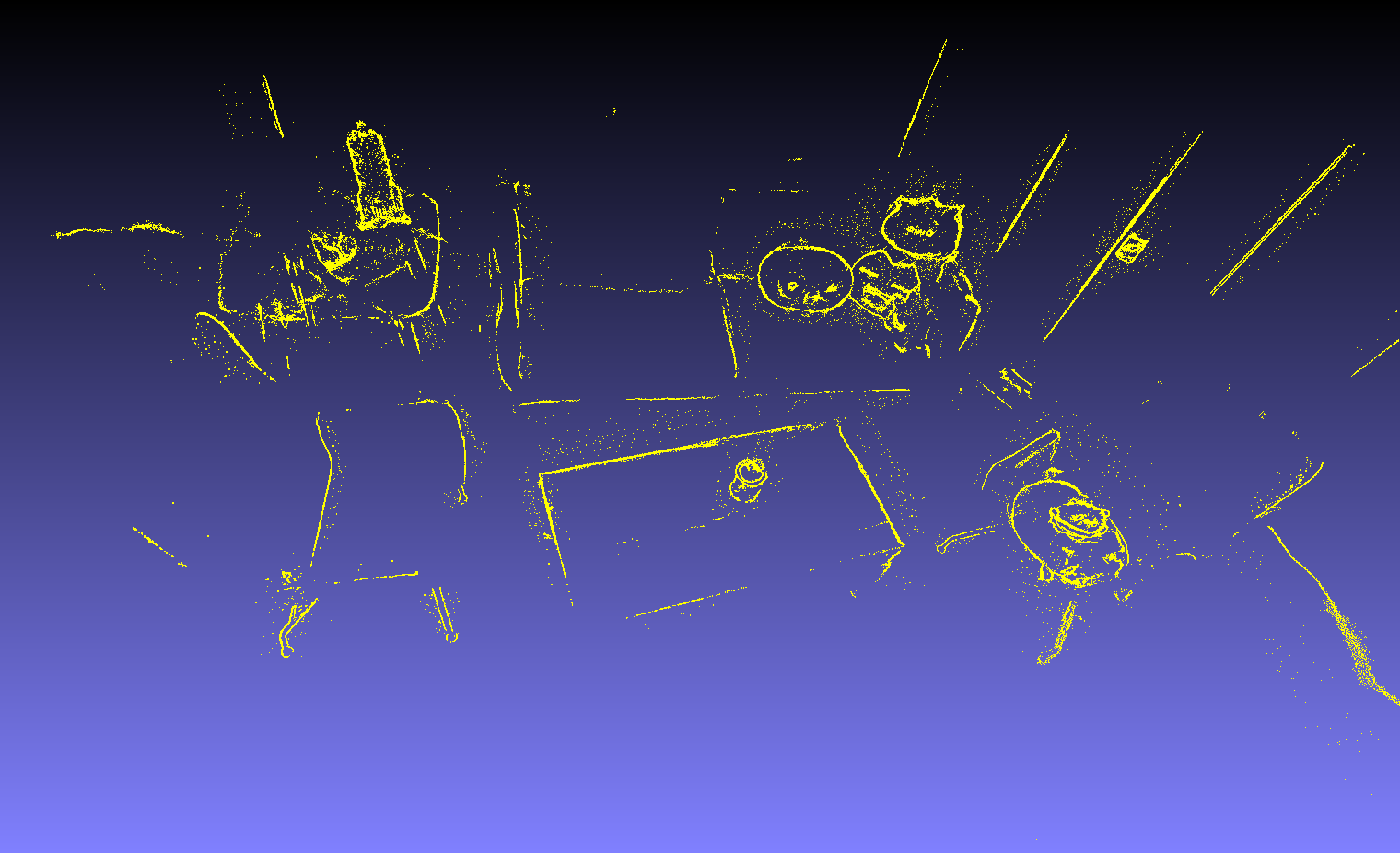}}
\subfigure[Global Semi-dense Map with 3D Gradient vectors]{\includegraphics[height = 0.21\textwidth]{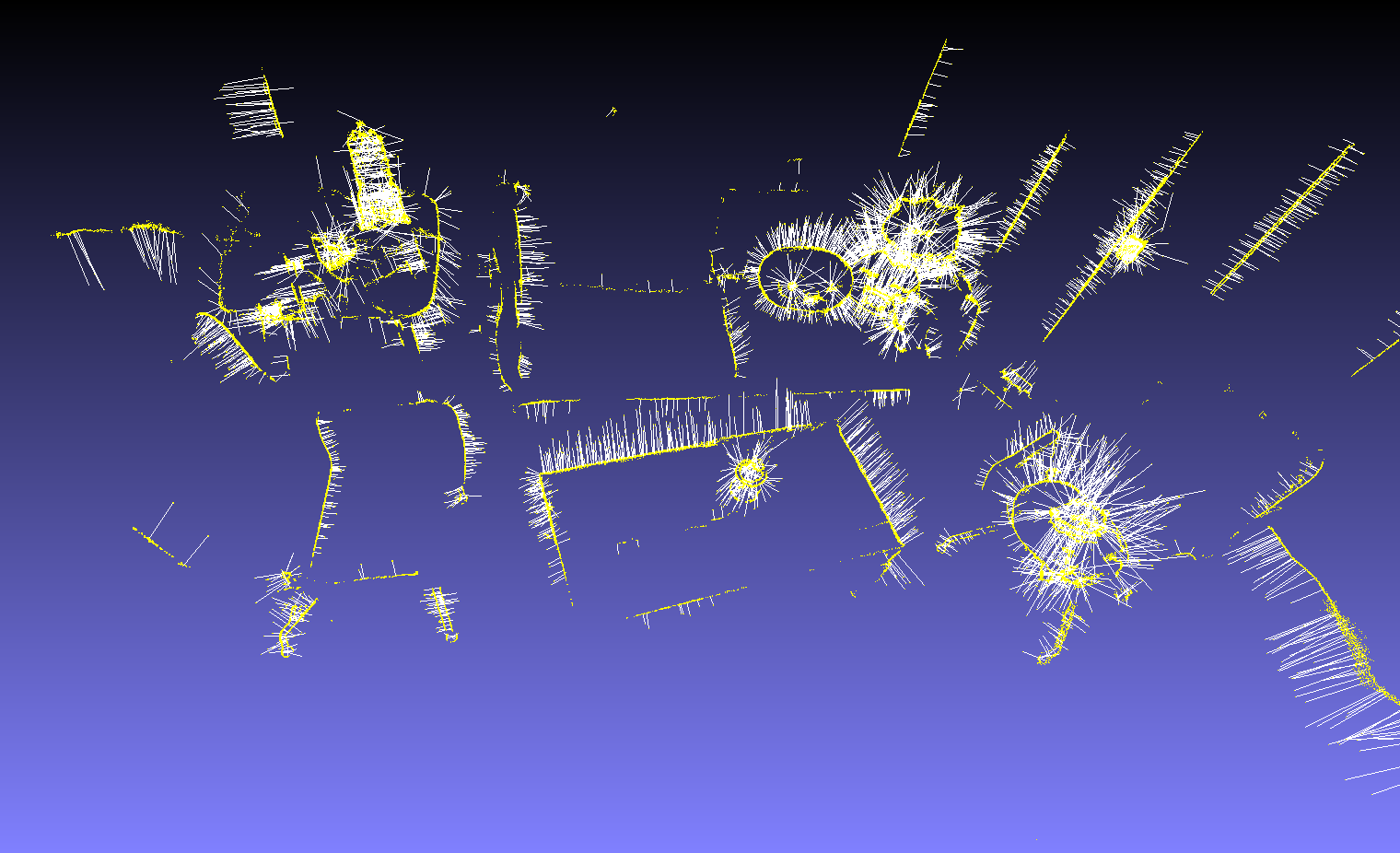}}
\vspace{-0.2cm}
\caption{\label{3D gradient}Visualization of a global semi-dense map (b) and its variant with 3D gradient vectors (c) for sequence sofa\_normal in \textit{VECtor}. An RGB image (a) is shown for reference.}
\end{figure*}

\subsubsection{Comparison against RGB-D camera-based solutions}
\label{robustness comparision}

With the primary objective to analyze robustness under challenging illumination conditions, we compare our method against two classical approaches that rely on RGB-D cameras or depth sensors, only. They are given by \textbf{KinectFusion}~\cite{newcombe2011kinectfusion} and \textbf{Canny-VO}~\cite{zhou2018canny}. We apply all methods to our self-collected datasets. We conduct three types of experiments, and all absolute trajectory errors (ATE) and relative pose errors (RPE) are summarized in Table~\ref{tab:all comparison}:
\begin{itemize}
    \item \textit{Variation of light conditions}: We apply all methods on a series of sequences with different illumination conditions denoted \textit{bright}, \textit{darkish}, \textit{dim}, \textit{dark} and \textit{high dynamic range (hdr)}. As summarized in Table~\ref{tab:MPL Availability}, both \textbf{Canny-DEVO} and \textbf{KinectFusion}~\cite{newcombe2011kinectfusion} are able to continuously track through all sequences, while \textbf{Canny-VO}~\cite{zhou2018canny} proves to be fragile when applied in poor illumination conditions. The reason is a lack of edge features caused by blur and poor contrast in dark scenarios. 
    \item \textit{Variation of motion characteristics}: We further evaluate the performance of all methods for different motion dynamics. The sequences are denoted \textit{fast}, \textit{mid}, or \textit{slow} to indicate the different camera dynamics. As can be observed in Table~\ref{tab:all comparison}, all methods have a remarkable ability to handle dynamic scenarios for standard depth camera frame rate.
    \item \textit{Variation of depth camera frame rate}: In order to analyse each method's ability to operate in an energy-saving mode, we finally test all methods for different depth camera frame rates between 30Hz and 1Hz in the \textit{table} environment and for three different camera dynamics. As indicated in Table \ref{tab:Frequency}, only our method is able to maintain stable tracking for all depth camera frame rates down to 1Hz. While accuracy decreases for more agile motion, it should be noted that the motion on these sequences is highly aggressive.
\end{itemize}

\subsection{Comparison of \textbf{Canny-EVT} against \textbf{Canny-VT}}
\label{Comparison Canny-EVT against Canny-VT}

The evaluation of \textbf{Canny-EVT} is conducted on our self-collected datasets, \revise{\textit{TUM}~\cite{sturm2012benchmark}}, and the \textit{VECtor} sequences. \revise{Here, since the \textit{TUM} datasets lack events, we employ rpg\_vid2e~\cite{gehrig2020video} to generate events using images}\revisetwo{, we adjust the contrast threshold between 0.1 to 0.5 according to the camera's speed in various data sequences to optimize the events generation results.} Our primary purpose is to demonstrate the advantages of \textbf{Canny-EVT} over \textbf{Canny-VT}, a regular vision-based alternative that performs similar, map-based semi-dense tracking. The experiments furthermore point out the advantage gained by including the polarity-aware \textbf{STSMs} and \textbf{ANNF}-based point culling.
\subsubsection{Details on implementation including the regular vision-based \textbf{Canny-VT}}
\label{Description of Canny-VT}

\begin{figure*}[t!]
\centering
\subfigure[Reprojection map with gradients and optical flow]{\label{repj_g_o}\includegraphics[height = 0.18\textwidth]{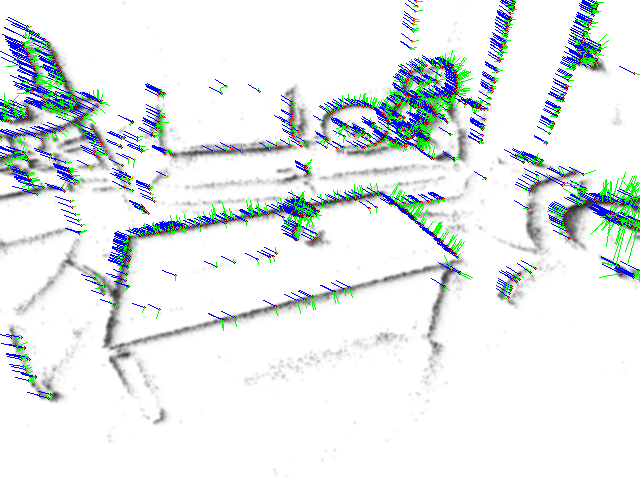}}
\subfigure[\textbf{TSM} with all events]{\label{stsm_all}\includegraphics[height = 0.18\textwidth]{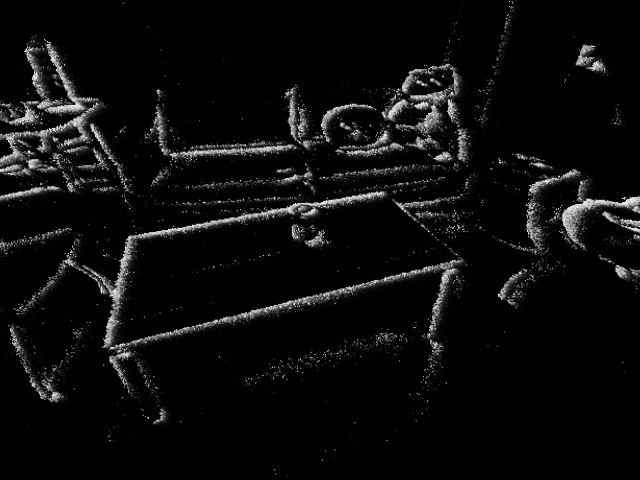}}
\subfigure[\textbf{TSM} with positive events]{\label{stsm_p}\includegraphics[height = 0.18\textwidth]{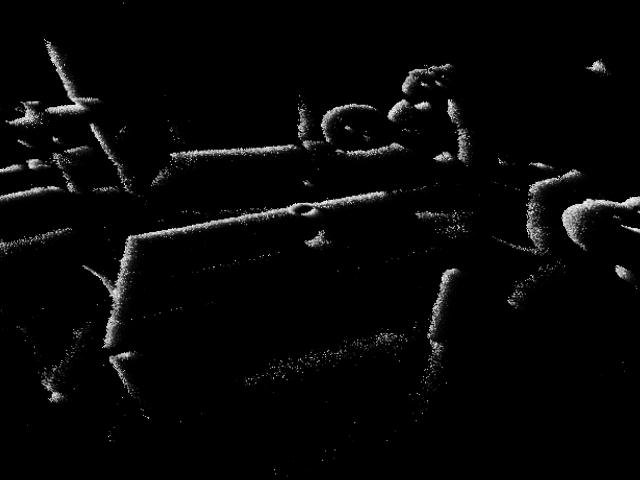}}
\subfigure[\textbf{TSM} with negative events]{\label{stsm_n}\includegraphics[height = 0.18\textwidth]{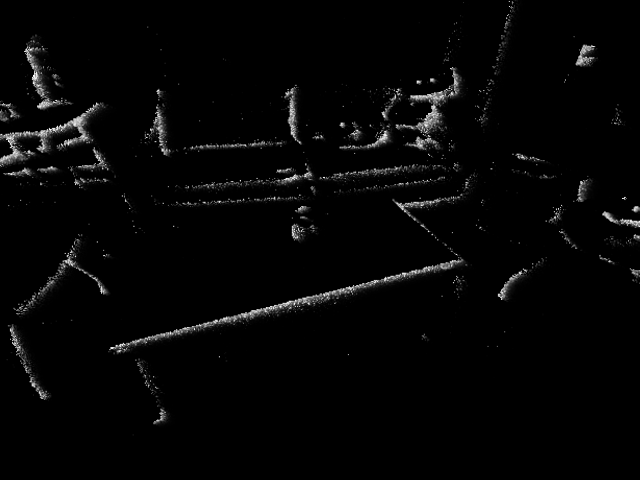}}
\caption{\label{STSM}Visualization of gradients and optical flow in reprojection map and \textbf{STSM}s. The green line is the projected gradient and the blue line is the predicted optical flow. If the gradient and the optical flow vectors have similar direction, the corresponding 3D point is predicted to trigger a positive event, and vice-versa. As an example, it is clear that the gradient and optical flow vectors have similar direction in the top-left corner of the table in Figure~\ref{repj_g_o}. As can be observed in the positive \textbf{TSM}~(\ref{stsm_p}), this part of the table indeed triggers positive events. Low potentials from negative events~(\ref{stsm_n}) are thus unable to attract these 3D points. Compared to the traditional \textbf{TSMs}~(\ref{stsm_all}), the registration with \textbf{STSM}s thus has improved ability to bridge large disparities.}
\end{figure*}

The global semi-dense point cloud is obtained by running released code~\cite{he2018incremental}. However, simple 3D points are insufficient in order to run our method. Polarity prediction requires the ability to predict the gradient and optical flow of every individual point. A key addition to the semi-dense mapping module hence consists of the 3D gradients which are obtained by projecting the corresponding image gradients onto local support planes. Figure~\ref{3D gradient} shows an example of added 3D gradient vectors. Optical flow is predicted by adding a constant velocity motion model. An example of projected predicted gradients and optical flow for every projected 3D point is indicated in Figure~\ref{STSM}.

In order to demonstrate the advantages of \textbf{Canny-EVT}, we implemented a regular vision-based pendant denoted \textbf{Canny-VT}. The latter is similar than \textbf{Canny-EVT}, except that edge measurements are extracted from regular images rather than from \textbf{TSM}s. Instead of the \textbf{TSM}-based potential field, we furthermore directly use a distance field derived from the edge maps extracted from the normal images, similar to one of the objectives persented in the \textbf{Canny-VO} paper~\cite{zhou2018canny}. The method hence performs 2D distance field based 3D-2D curve registration, and residuals are calculated by directly sampling the distance field at the 3D point reprojection locations. In order to properly compare \textbf{Canny-VT} against \textbf{Canny-EVT}, both methods make use of the exact same global semi-dense point cloud, and both methods implement the occlusion handling strategy obtained by the addition of ANNFs. For both methods, the initial pose is predicted by a constant velocity motion model, followed by projection of the global semi-dense map to the current frame.

\begin{table*}[h!]
  \centering
\captionsetup{justification=centering}
  \caption{
  RPE and ATE on self-collected, \textit{VECtor} and \textit{TUM} datasets for \textbf{Canny-VT}, \textbf{Canny-EVT-{}-} and \textbf{Canny-EVT}.\\ $\left[\mathbf{R}_\text{\upshape rpe}\text{\upshape : °/s},\mathbf{t}_\text{\upshape rpe}\text{\upshape : cm/s},\mathbf{R}_\text{\upshape ate}\text{\upshape : °}, \mathbf{t}_\text{\upshape ate}\text{\upshape : cm}\right]$}
  \label{tab:Ablation Study table}
    \begin{tabular}{llcccccccccccccc}
    \toprule
          &       & \multicolumn{4}{c}{\textbf{Canny-VT}} &       & \multicolumn{4}{c}{\textbf{Canny-EVT-{}-}} &       & \multicolumn{4}{c}{\textbf{Canny-EVT}} \\
\cmidrule{3-6}\cmidrule{8-11}\cmidrule{13-16}    \textit{Dataset} & \textit{Sequence} & $\mathbf{R}_\text{rpe}$  & $\mathbf{t}_\text{rpe}$ & $\mathbf{R}_\text{ate}$ & $\mathbf{t}_\text{ate}$ &       & $\mathbf{R}_\text{rpe}$  & $\mathbf{t}_\text{rpe}$ & $\mathbf{R}_\text{ate}$ & $\mathbf{t}_\text{ate}$ &       & $\mathbf{R}_\text{rpe}$  & $\mathbf{t}_\text{rpe}$ & $\mathbf{R}_\text{ate}$ & $\mathbf{t}_\text{ate}$ \\
\cmidrule{1-6}\cmidrule{8-11}\cmidrule{13-16}    \multirow{11}[2]{*}{\textit{Self-collected}} & \textit{table\_bright\_slow} & 1.06  & 0.73  & \textbf{2.45} & 1.11  &       & \textbf{0.74} & 0.65  & 3.69  & \textbf{0.71} &       & 0.77  & \textbf{0.61} & 2.91  & 0.85 \\
          & \textit{table\_bright\_mid} & 1.31  & 0.92  & \textbf{1.79} & 1.42  &       & \textbf{1.15} & \textbf{0.91} & 2.23  & \textbf{1.19} &       & 1.23  & 0.94  & 2.44  & 1.25 \\
          & \textit{table\_bright\_fast} & 1.72  & \textbf{1.66} & \textbf{2.51} & \textbf{2.16} &       & 1.54  & 2.00  & 3.07  & 2.49  &       & \textbf{1.50} & 1.87  & 2.98  & 2.40 \\
          & \textit{sofa\_bright\_slow} & 3.98  & 3.48  & 6.19  & 4.88  &       & \textbf{1.14} & 1.34  & \textbf{3.05} & 1.53  &       & 1.15  & \textbf{1.17} & 3.28  & \textbf{1.49} \\
          & \textit{cali\_bright\_slow} & \textbf{0.46} & \textbf{0.57} & \textbf{2.19} & \textbf{0.78} &       & 1.27  & 2.00  & 5.98  & 2.58  &       & 1.30  & 1.83  & 5.89  & 2.53 \\
          & \textit{cali\_bright\_mid} & 4.18  & 1.94  & 8.89  & 2.25  &       & \textbf{1.52} & 1.33  & \textbf{2.49} & \textbf{2.23} &       & 1.58  & \textbf{1.26} & 2.66  & 2.25 \\
          & \textit{cali\_bright\_fast} & -     & -     & -     & -     &       & 4.31  & 8.43  & \textbf{45.24} & 20.78 &       & \textbf{4.21} & \textbf{7.77} & 76.88 & \textbf{18.82} \\
          & \textit{table\_normal} & 3.35  & 2.73  & 8.04  & 2.42  &       & 0.89  & 1.39  & 3.58  & 1.05  &       & \textbf{0.82} & \textbf{1.22} & \textbf{3.42} & \textbf{0.95} \\
          & \textit{table\_aggressive} & -     & -     & -     & -     &       & 3.56  & 3.08  & 6.42  & \textbf{4.60} &       & \textbf{3.04} & \textbf{2.58} & \textbf{6.29} & 4.67 \\
          & \textit{sofa\_normal} & 3.14  & 1.84  & 3.24  & 1.59  &       & 8.57  & 11.60 & 8.80  & 11.45 &       & \textbf{1.00} & \textbf{0.89} & \textbf{2.32} & \textbf{0.76} \\
          & \textit{desk\_normal} & 2.58  & 2.20  & 3.02  & 2.10  &       & 0.95  & 1.11  & 2.43  & 0.82  &       & \textbf{0.85} & \textbf{0.90} & \textbf{2.35} & \textbf{0.72} \\
\cmidrule{1-6}\cmidrule{8-11}\cmidrule{13-16}    \multirow{7}[2]{*}{\textit{VECtor}} & \textit{board\_slow} & 0.42  & \textbf{0.85} & 1.97  & \textbf{0.64} &       & 0.31  & 1.45  & \textbf{1.59} & 1.20  &       & \textbf{0.29} & 1.45  & 1.60  & 1.19 \\
          & \textit{mountain\_fast} & -     & -     & -     & -     &       & \textbf{2.68} & \textbf{2.73} & \textbf{2.50} & \textbf{1.82} &       & 3.59  & 2.95  & 4.24  & 2.14 \\
          & \textit{desk\_fast} & -     & -     & -     & -     &       & -     & -     & -     & -     &       & \textbf{4.45} & \textbf{5.42} & \textbf{6.70} & \textbf{5.80} \\
          & \textit{robot\_fast} & -     & -     & -     & -     &       & -     & -     & -     & -     &       & \textbf{6.27} & \textbf{11.70} & \textbf{6.05} & \textbf{9.07} \\
          & \textit{robot\_normal} & 1.41  & \textbf{1.52} & \textbf{1.32} & \textbf{0.92} &       & 1.29  & 2.13  & 2.43  & 1.58  &       & \textbf{1.26} & 2.06  & 2.46  & 1.58 \\
          & \textit{sofa\_fast} & -     & -     & -     & -     &       & \textbf{2.40} & 2.70  & 4.53  & 1.99  &       & 2.54  & \textbf{1.35} & \textbf{2.84} & \textbf{1.29} \\
          & \textit{sofa\_normal} & 5.01  & 3.93  & 5.80  & 4.00  &       & 0.97  & 1.04  & \textbf{1.13} & 1.09  &       & \textbf{0.96} & \textbf{0.95} & 1.17  & \textbf{1.06} \\
          \cmidrule{1-6}\cmidrule{8-11}\cmidrule{13-16}
          \multirow{4}[0]{*}{\revise{\textit{TUM}}} & \revise{\textit{fr2\_desk}} & \revise{0.015} & \revise{0.98}  & \revise{0.019}  & \revise{4.35}  &       & \revise{0.014} & \revise{0.92}  & \revise{0.015}  & \revise{1.39}  &       & \revise{\textbf{0.013}} & \revise{\textbf{0.74}} & \revise{\textbf{0.011}} & \revise{\textbf{0.79}} \\
          & \revise{\textit{fr2\_rpy}} & \revise{-} & \revise{-} & \revise{-} & \revise{-} &       & \revise{0.004} & \revise{0.36}  & \revise{0.002} & \revise{10.39} &       & \revise{\textbf{0.003}} & \revise{\textbf{0.27}} & \revise{\textbf{0.002}} & \revise{\textbf{6.69}} \\
          & \revise{\textit{fr3\_long\_office}} & \revise{0.026}  & \revise{1.86}  & \revise{0.019}  & \revise{1.56}  &       & \revise{0.083} & \revise{1.21}  & \revise{0.082} & \revise{1.20}  &       & \revise{\textbf{0.077}} & \revise{\textbf{0.47}} & \revise{\textbf{0.061}} & \revise{\textbf{0.60}}
          \\
          & \revise{\textit{fr1\_xyz}} & \revise{0.058}  & \revise{1.91}  & \revise{0.032}  & \revise{7.95}  &       & \revise{0.038}  & \revise{1.63}  & \revise{0.024}  & \revise{11.28} &       & \revise{\textbf{0.010}} & \revise{\textbf{0.46}} & \revise{\textbf{0.006}} & \revise{\textbf{0.95}}
          \\
    \bottomrule
    \end{tabular}%
\end{table*}%

\begin{figure*}[h]
\centering
\subfigure[Scene]{\label{rgb_occluded}\includegraphics[height = 0.187\textwidth]{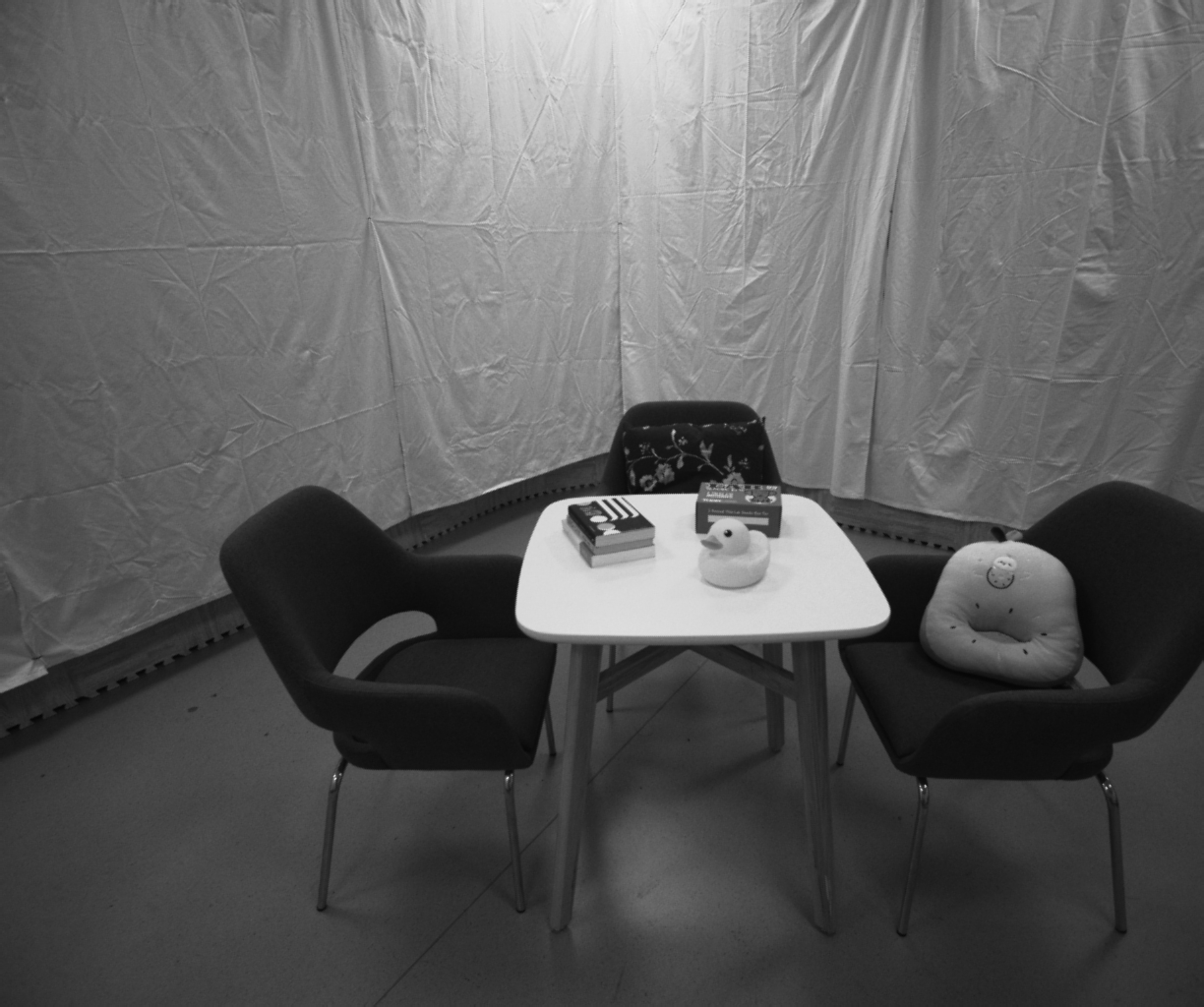}}
\subfigure[Time Surface Map]{\label{ts_occluded}\includegraphics[height = 0.187\textwidth]{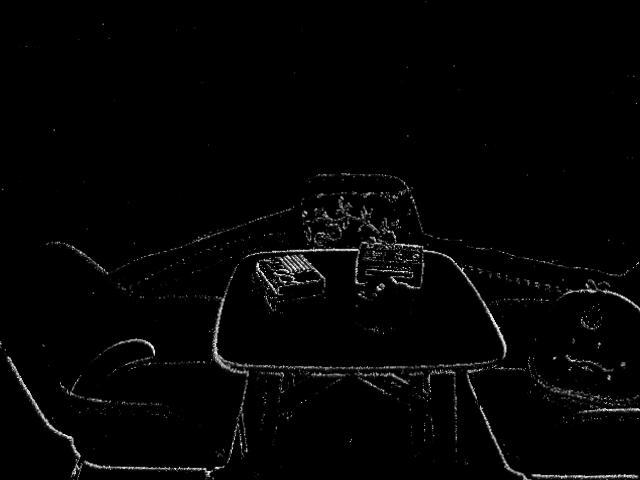}}
\subfigure[Reprojection Map by \textbf{Canny-EVT-{}-}]{\label{repj_occluded}\includegraphics[height = 0.187\textwidth]{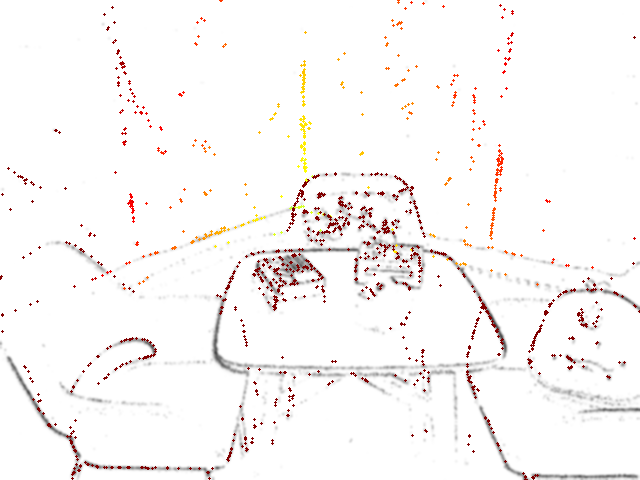}}
\subfigure[Reprojection Map by \textbf{Canny-EVT}]{\label{repj_without_occluded}\includegraphics[height = 0.187\textwidth]{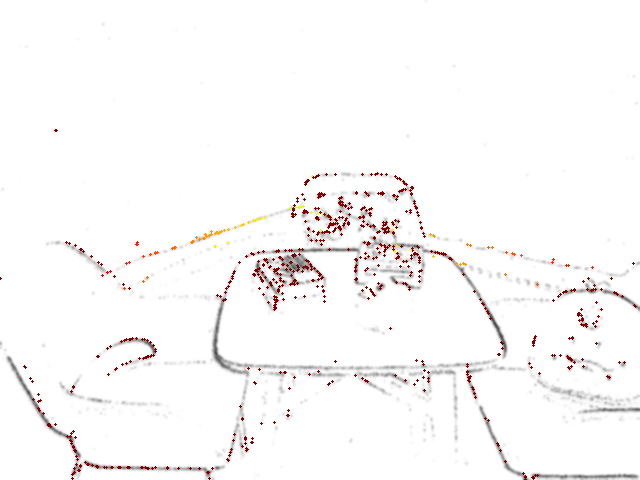}}
\caption{\label{Visualizaion of Occluded points}Visualization of an image (a) and \textbf{TSM} (b) from sequence \textit{table\_bright\_slow} from our self-collected datasets. On the right, we see corresponding reprojections into a nearby frame for \textbf{Canny-EVT-{}-} (c) and \textbf{Canny-EVT} (d). The coloring indicates the depth of each point. As can be observed, ANNF-based point culling helps to remove many outlier associations.}
\vspace{-0.4cm}
\end{figure*}



\begin{table*}[h]
\vspace{0.5cm}
  \centering
\captionsetup{justification=centering}  
  \caption{RPE and ATE on self-collected datasets for different \textbf{TSM} rates.\\ $\left[\mathbf{R}_\text{\upshape rpe}\text{\upshape : °/s},\mathbf{t}_\text{\upshape rpe}\text{\upshape : cm/s},\mathbf{R}_\text{\upshape ate}\text{\upshape : °}, \mathbf{t}_\text{\upshape ate}\text{\upshape : cm}\right]$}
  \label{tab:Ablation Study_ours}
    \begin{tabular}{lcccccccccc}
    \toprule
          &       & \multicolumn{4}{c}{\textbf{Canny-EVT-{}-}} &       & \multicolumn{4}{c}{\textbf{Canny-EVT}} \\
\cmidrule{3-6}\cmidrule{8-11}    \textit{Sequence} & \multicolumn{1}{l}{\textit{Frequency}} & $\mathbf{R}_\text{rpe}$  & $\mathbf{t}_\text{rpe}$ & $\mathbf{R}_\text{ate}$ & $\mathbf{t}_\text{ate}$ &       & $\mathbf{R}_\text{rpe}$  & $\mathbf{t}_\text{rpe}$ & $\mathbf{R}_\text{ate}$ & $\mathbf{t}_\text{ate}$ \\
\cmidrule{1-6}\cmidrule{8-11}    \multirow{3}[2]{*}{\textit{cali\_bright\_slow}} & \textit{300} & \textbf{1.27} & 2.00  & 5.98  & 2.58  &       & 1.30  & \textbf{1.83} & \textbf{5.89} & \textbf{2.53} \\
          & \textit{150} & 2.68  & 4.57  & 14.73 & 9.40  &       & \textbf{1.07} & \textbf{2.52} & \textbf{6.76} & \textbf{3.40} \\
          & \textit{100} & 1.84  & 3.32  & \textbf{7.44} & 7.58  &       & \textbf{1.76} & \textbf{2.81} & 14.35 & \textbf{6.12} \\
\cmidrule{1-6}\cmidrule{8-11}    \multirow{3}[2]{*}{\textit{cali\_bright\_mid}} & \textit{300} & \textbf{1.52} & \textbf{1.33} & \textbf{2.49} & \textbf{2.23} &       & 1.58  & \textbf{1.26} & 2.66  & 2.25 \\
          & \textit{150} & \textbf{1.73} & \textbf{1.57} & 3.49  & \textbf{2.38} &       & 1.74  & \textbf{1.56} & \textbf{3.36} & 2.48 \\
          & \textit{100} & -     & -     & -     & -     &       & \textbf{2.08} & \textbf{2.17} & \textbf{4.93} & \textbf{2.96} \\
\cmidrule{1-6}\cmidrule{8-11}    \textit{cali\_bright\_fast} & \textit{300} & 4.31  & 8.43  & \textbf{45.24} & 20.78 &       & \textbf{4.21} & \textbf{7.77} & 76.88 & \textbf{18.82} \\
\cmidrule{1-6}\cmidrule{8-11}    \multirow{3}[2]{*}{\textit{table\_bright\_slow}} & \textit{300} & \textbf{0.75} & 0.70  & 4.15  & \textbf{0.73} &       & \textbf{0.75} & \textbf{0.64} & \textbf{3.03} & 0.87 \\
          & \textit{150} & \textbf{0.74} & 0.65  & 3.69  & \textbf{0.71} &       & 0.77  & \textbf{0.61} & \textbf{2.91} & 0.85 \\
          & \textit{100} & 1.13  & 2.72  & 10.30 & 3.01  &       & \textbf{0.98} & \textbf{1.01} & \textbf{4.10} & \textbf{1.13} \\
\cmidrule{1-6}\cmidrule{8-11}    \multirow{3}[2]{*}{\textit{table\_bright\_mid}} & \textit{300} & \textbf{1.15} & \textbf{0.91} & \textbf{2.23} & \textbf{1.19} &       & 1.42  & 0.94  & 2.93  & 1.33 \\
          & \textit{150} & \textbf{1.19} & 1.55  & \textbf{2.44} & 2.74  &       & 1.23  & \textbf{0.94} & \textbf{2.44} & \textbf{1.25} \\
          & \textit{100} & 1.74  & 2.19  & 3.79  & 3.45  &       & \textbf{1.60} & \textbf{1.17} & \textbf{2.94} & \textbf{1.46} \\
\cmidrule{1-6}\cmidrule{8-11}    \multirow{3}[2]{*}{\textit{table\_bright\_fast}} & \textit{300} & 1.54  & 2.00  & 3.07  & 2.49  &       & \textbf{1.50} & \textbf{1.87} & \textbf{2.98} & \textbf{2.40} \\
          & \textit{150} & 1.65  & 1.84  & 3.22  & 2.50  &       & \textbf{1.59} & \textbf{1.71} & \textbf{3.07} & \textbf{2.39} \\
          & \textit{100} & \textbf{2.95} & \textbf{4.25} & \textbf{9.48} & \textbf{6.20} &       & 2.97  & 5.09  & 10.37 & 7.02 \\
\cmidrule{1-6}\cmidrule{8-11}    \multirow{2}[2]{*}{\textit{sofa\_bright\_slow}} & \textit{300} & \textbf{1.14} & 1.34  & \textbf{3.05} & 1.53  &       & 1.15  & \textbf{1.17} & 3.28  & \textbf{1.49} \\
          & \textit{150} & \textbf{1.31} & \textbf{1.48} & \textbf{4.18} & \textbf{1.77} &       & 1.35  & 1.51  & 4.30  & 1.88 \\
\cmidrule{1-6}\cmidrule{8-11}    \multirow{3}[2]{*}{\textit{table\_normal}} & \textit{300} & 0.91  & 1.52  & 3.76  & 1.13  &       & \textbf{0.83} & \textbf{1.29} & \textbf{3.42} & \textbf{0.96} \\
          & \textit{150} & 0.89  & 1.39  & 3.58  & 1.05  &       & \textbf{0.82} & \textbf{1.22} & \textbf{3.42} & \textbf{0.95} \\
          & \textit{100} & \textbf{0.79} & 1.55  & 3.96  & 1.19  &       & 0.82  & \textbf{1.31} & \textbf{3.81} & \textbf{0.99} \\
\cmidrule{1-6}\cmidrule{8-11}    \multirow{2}[2]{*}{\textit{table\_aggressive}} & \textit{300} & 3.56  & 3.08  & 6.42  & \textbf{4.60} &       & \textbf{3.04} & \textbf{2.58} & \textbf{6.29} & 4.67 \\
          & \textit{150} & -     & -     & -     & -     &       & \textbf{5.11} & \textbf{5.15} & \textbf{12.90} & \textbf{7.22} \\
\cmidrule{1-6}\cmidrule{8-11}    \multirow{3}[2]{*}{\textit{sofa\_normal}} & \textit{300} & 9.87  & 49.70 & 16.44 & 48.74 &       & \textbf{1.00} & \textbf{0.89} & \textbf{2.32} & \textbf{0.76} \\
          & \textit{150} & 8.57  & 11.60 & 8.80  & 11.45 &       & \textbf{1.06} & \textbf{1.13} & \textbf{2.39} & \textbf{1.04} \\
          & \textit{100} & -     & -     & -     & -     &       & \textbf{1.23} & \textbf{1.46} & \textbf{2.55} & \textbf{1.43} \\
\cmidrule{1-6}\cmidrule{8-11}    \multirow{3}[2]{*}{\textit{desk\_normal}} & \textit{300} & 0.95  & 1.11  & 2.43  & 0.82  &       & \textbf{0.85} & \textbf{0.90} & \textbf{2.35} & \textbf{0.72} \\
          & \textit{150} & 16.93 & 11.97 & 23.02 & 13.89 &       & \textbf{0.91} & \textbf{0.98} & \textbf{2.33} & \textbf{0.74} \\
          & \textit{100} & -     & -     & -     & -     &       & \textbf{1.00} & \textbf{0.99} & \textbf{2.35} & \textbf{0.83} \\
    \bottomrule
    \end{tabular}%
    \vspace{0.6cm}
\end{table*}%

\subsubsection{Results}
\label{Results}

Table \ref{tab:Ablation Study table} shows the evaluation results obtained over all sequences we mentioned before. The best results per sequence are highlighted in bold. It is easy to observe that ATE and RPE errors are generally lower for \textbf{Canny-EVT} than they are for \textbf{Canny-VT}, especially for fast motion.  \textbf{Canny-VT} fails to handle highly dynamic situations for two reasons. First, canny edge detection fails to detect clear edges in RGB images with motion blur. Second, our normal camera does not have sufficiently high frame rate, which causes large disparities and thus convergence to wrong local minima in highly dynamic scenarios.


We furthermore compare the  results of \textbf{Canny-EVT} against a variant for which polarity-aware registration and occlusion reasoning is switched off. The simplified alternative is denoted \textbf{Canny-EVT-{}-}. As can be observed, results for \textbf{Canny-EVT} are typically better than for \textbf{Canny-EVT-{}-}, especially in the scenes \textit{sofa} and \textit{desk}. 3D-2D edge alignment relies on clear \textbf{TSM}s and good initial poses. Complex scenes with more details will trigger many events. Wrong assignment to edges that have incompatible polarity thus become more likely. As explained in Section \ref{polarity-aware registration by STSMs}, \textbf{STSM}s can alleviate this issue by dividing the assignments of the reprojected 3D points into two distinct \textbf{TSM}s. Higher accuracy is furthermore supported by the \textbf{ANNF}-supported occlusion reasoning, the effect of which is visualized in Figure~\ref{Visualizaion of Occluded points}. In general, \textbf{Canny-EVT} and \textbf{Canny-EVT-{}-} both can work better than \textbf{Canny-VT} in more dynamic situations, and \textbf{Canny-EVT} works outstandingly  well even in scenes with more complex texture. \textbf{Canny-EVT-{}-} only shows better performance when the event camera is shaken dramatically as it is the case in sequences \textit{table} and \textit{calibration} of our self-collected datasets and the sequence \textit{mountain\_fast} of \textit{VECtor}. Aggressive, jerky motion may indeed affect the accuracy of optical flow predictions and thereby make the prediction of the polarity inaccurate.

\begin{table*}[h!]
  \centering
\captionsetup{justification=centering}  
  \caption{RPE and ATE on \textit{VECtor} dataset for different \textbf{TSM} rates.\\ $\left[\mathbf{R}_\text{\upshape rpe}\text{\upshape : °/s},\mathbf{t}_\text{\upshape rpe}\text{\upshape : cm/s},\mathbf{R}_\text{\upshape ate}\text{\upshape : °}, \mathbf{t}_\text{\upshape ate}\text{\upshape : cm}\right]$}
  \label{tab:Ablation Study_VECtor}
    \begin{tabular}{lcccccccccc}
    \toprule
          &       & \multicolumn{4}{c}{\textbf{Canny-EVT-{}-}} &       & \multicolumn{4}{c}{\textbf{Canny-EVT}} \\
\cmidrule{3-6}\cmidrule{8-11}    \textit{Sequence} & \multicolumn{1}{l}{\textit{Frequency}} & $\mathbf{R}_\text{rpe}$  & $\mathbf{t}_\text{rpe}$ & $\mathbf{R}_\text{ate}$ & $\mathbf{t}_\text{ate}$ &       & $\mathbf{R}_\text{rpe}$  & $\mathbf{t}_\text{rpe}$ & $\mathbf{R}_\text{ate}$ & $\mathbf{t}_\text{ate}$ \\
\cmidrule{1-6}\cmidrule{8-11}    \multirow{3}[2]{*}{\textit{sofa\_normal}} & \textit{300} & 0.97  & \textbf{1.04} & \textbf{1.13} & 1.09  &       & \textbf{0.96} & \textbf{0.95} & 1.17  & \textbf{1.06} \\
          & \textit{150} & \textbf{1.07} & 1.15  & \textbf{1.24} & 1.20  &       & \textbf{1.07} & \textbf{1.10} & 1.27  & \textbf{1.19} \\
          & \textit{100} & 1.49  & 1.91  & 1.78  & 2.00  &       & \textbf{1.19} & \textbf{1.20} & \textbf{1.45} & \textbf{1.33} \\
\cmidrule{1-6}\cmidrule{8-11}    \multirow{3}[2]{*}{\textit{robot\_normal}} & \textit{300} & 1.29  & \textbf{2.13} & \textbf{2.43} & \textbf{1.58} &       & \textbf{1.26} & \textbf{2.06} & 2.46  & \textbf{1.58} \\
          & \textit{150} & 1.47  & \textbf{2.34} & 2.85  & \textbf{1.67} &       & \textbf{1.37} & 2.38  & \textbf{2.70} & 1.69 \\
          & \textit{100} & 1.71  & 2.82  & 3.13  & 2.03  &       & \textbf{1.52} & \textbf{2.71} & \textbf{3.04} & \textbf{1.79} \\
\cmidrule{1-6}\cmidrule{8-11}    \multirow{3}[2]{*}{\textit{board\_slow}} & \textit{300} & 0.35  & 1.51  & \textbf{1.56} & 1.24  &       & \textbf{0.32} & \textbf{1.50} & \textbf{1.56} & \textbf{1.22} \\
          & \textit{150} & 0.31  & \textbf{1.45} & \textbf{1.59} & 1.20  &       & \textbf{0.29} & \textbf{1.45} & 1.60  & \textbf{1.19} \\
          & \textit{100} & 0.32  & 1.54  & 1.76  & 1.26  &       & \textbf{0.29} & \textbf{1.52} & \textbf{1.69} & \textbf{1.24} \\
\cmidrule{1-6}\cmidrule{8-11}    \textit{robot\_fast} & \textit{300} & -     & -     & -     & -     &       & \textbf{6.27} & \textbf{11.70} & \textbf{6.05} & \textbf{9.07} \\
\cmidrule{1-6}\cmidrule{8-11}    \multirow{2}[2]{*}{\textit{mountain\_fast}} & \textit{450} & \textbf{2.68} & \textbf{2.73} & \textbf{2.50} & \textbf{1.82} &       & 3.59  & 2.95  & 4.24  & 2.14 \\
          & \textit{300} & -     & -     & -     & -     &       & \textbf{3.58} & \textbf{3.21} & \textbf{4.79} & \textbf{2.55} \\
\cmidrule{1-6}\cmidrule{8-11}    \multirow{3}[2]{*}{\textit{sofa\_fast}} & \textit{450} & \textbf{2.40} & 2.70  & 4.53  & 1.99  &       & 2.54  & \textbf{1.35} & \textbf{2.84} & \textbf{1.29} \\
          & \textit{300} & -     & -     & -     & -     &       & \textbf{2.57} & \textbf{1.50} & \textbf{3.14} & \textbf{1.43} \\
          & \textit{150} & -     & -     & -     & -     &       & \textbf{3.18} & \textbf{2.45} & \textbf{6.74} & \textbf{2.16} \\
\cmidrule{1-6}\cmidrule{8-11}    \textit{desk\_fast} & \textit{300} & -     & -     & -     & -     &       & \textbf{4.45} & \textbf{5.42} & \textbf{6.70} & \textbf{5.80} \\
    \bottomrule
    \end{tabular}%
    \vspace{0.5cm}
\end{table*}%

\begin{table*}[h]
  \centering
  \caption{\revise{RPE and ATE on \textit{TUM} dataset for different \textbf{TSM} rates.\\ $\left[\mathbf{R}_\text{\upshape rpe}\text{\upshape : °/s},\mathbf{t}_\text{\upshape rpe}\text{\upshape : cm/s},\mathbf{R}_\text{\upshape ate}\text{\upshape : °}, \mathbf{t}_\text{\upshape ate}\text{\upshape : cm}\right]$}}
  \revise{
    \begin{tabular}{lcccccccccc}
    \toprule
          &       & \multicolumn{4}{c}{\textbf{Canny-EVT-{}-}} &       & \multicolumn{4}{c}{\textbf{Canny-EVT}} \\
\cmidrule{3-6}\cmidrule{8-11}    \textit{Sequence} & \textit{Frequency} & $\mathbf{R}_\text{rpe}$  & $\mathbf{t}_\text{rpe}$ & $\mathbf{R}_\text{ate}$  & $\mathbf{t}_\text{ate}$ &       & $\mathbf{R}_\text{rpe}$  & $\mathbf{t}_\text{rpe}$ & $\mathbf{R}_\text{ate}$  & $\mathbf{t}_\text{ate}$ \\
    \midrule
    \multirow{3}[2]{*}{\textit{fr2\_desk}} & \textit{300} & 0.01  & 0.34  & 0.01  & 1.70  &       & \textbf{0.00} & \textbf{0.30} & \textbf{0.00} & \textbf{0.59} \\
          & \textit{150} & 0.01  & 0.53  & 0.01  & 1.91  &       & \textbf{0.01} & \textbf{0.36} & \textbf{0.00} & \textbf{0.53} \\
          & \textit{100} & 0.01  & 0.92  & 0.02  & 1.39  &       & \textbf{0.01} & \textbf{0.74} & \textbf{0.01} & \textbf{0.79} \\
\cmidrule{1-6}\cmidrule{8-11}    \multirow{3}[2]{*}{\textit{fr2\_rpy}} & \textit{300} & 0.00  & 0.40  & 0.00  & 7.96  &       & \textbf{0.00} & \textbf{0.29} & \textbf{0.00} & \textbf{4.98} \\
          & \textit{150} & 0.00  & 0.36  & 0.00  & 10.39 &       & \textbf{0.00} & \textbf{0.27} & \textbf{0.00} & \textbf{6.69} \\
          & \textit{100} & 0.00  & 0.32  & 0.00  & 13.25 &       & \textbf{0.00} & \textbf{0.30} & \textbf{0.00} & \textbf{11.46} \\
\cmidrule{1-6}\cmidrule{8-11}    \multirow{3}[2]{*}{\textit{fr3\_long\_office}} & \textit{300} & 0.01  & 1.50  & \textbf{0.01} & 1.29  &       & \textbf{0.01} & \textbf{1.46} & 0.01  & \textbf{1.20} \\
          & \textit{150} & 0.01  & 1.21  & 0.01  & 1.20  &       & \textbf{0.01} & \textbf{0.47} & \textbf{0.01} & \textbf{0.60} \\
          & \textit{100} & \textbf{0.01} & 1.12  & 0.01  & 1.18  &       & 0.01  & \textbf{0.63} & \textbf{0.01} & \textbf{0.71} \\
\cmidrule{1-6}\cmidrule{8-11}    \multirow{3}[2]{*}{\textit{fr1\_xyz}} & \textit{300} & 0.04  & 1.63  & 0.02  & 11.28 &       & \textbf{0.01} & \textbf{0.46} & \textbf{0.01} & \textbf{0.95} \\
          & \textit{150} & 0.07  & 2.91  & 0.03  & 45.06 &       & \textbf{0.04} & \textbf{1.87} & \textbf{0.03} & \textbf{19.15} \\
          & \textit{100} & 0.08  & 3.60  & 0.04  & 58.61 &       & \textbf{0.06} & \textbf{2.57} & \textbf{0.03} & \textbf{38.08} \\
    \bottomrule
    \end{tabular}%
    }
  \label{tab:tum_frequency}%
  \vspace{-0.2cm}
\end{table*}%

\subsubsection{Results for Different Frequency}
\label{results for different frequency}

To further evaluate the advantage of polarity-aware registration by \textbf{STSM}s, we compare \textbf{Canny-EVT-{}-} and \textbf{Canny-EVT} for different \textbf{TSM} frame rates. It is intuitively clear that lower framerates will lead to larger disparities, and thereby grow the risk of convergence into wrong local minima. The addition of \textbf{STSM}s again helps to enlarge convergence basins, and thereby ensure correct convergence. Related evaluation results are given in Tables \ref{tab:Ablation Study_ours}, \ref{tab:Ablation Study_VECtor} and \ref{tab:tum_frequency}. As can be observed, the \textbf{STSM}s-based tracker generally works better than the \textbf{TSM}-based tracker, especially for low frame rates. As can be further observed in Table \ref{tab:Ablation Study_ours}, \textbf{Canny-EVT-{}-} fails to complete sequences \textit{cali\_bright\_mid}, \textit{sofa\_normal} and \textit{desk\_normal} when the frequency is reduced to less than 100Hz, whereas \textbf{Canny-EVT} continues to work well. A similar situation can be observed in Table \ref{tab:Ablation Study_VECtor} on sequences \textit{mountain\_fast} and \textit{sofa\_desk}. 

\begin{figure}[b!]
\vspace{-0.3cm}
\centering
\subfigure[Scene in the dark]{\includegraphics[width=4.3cm]{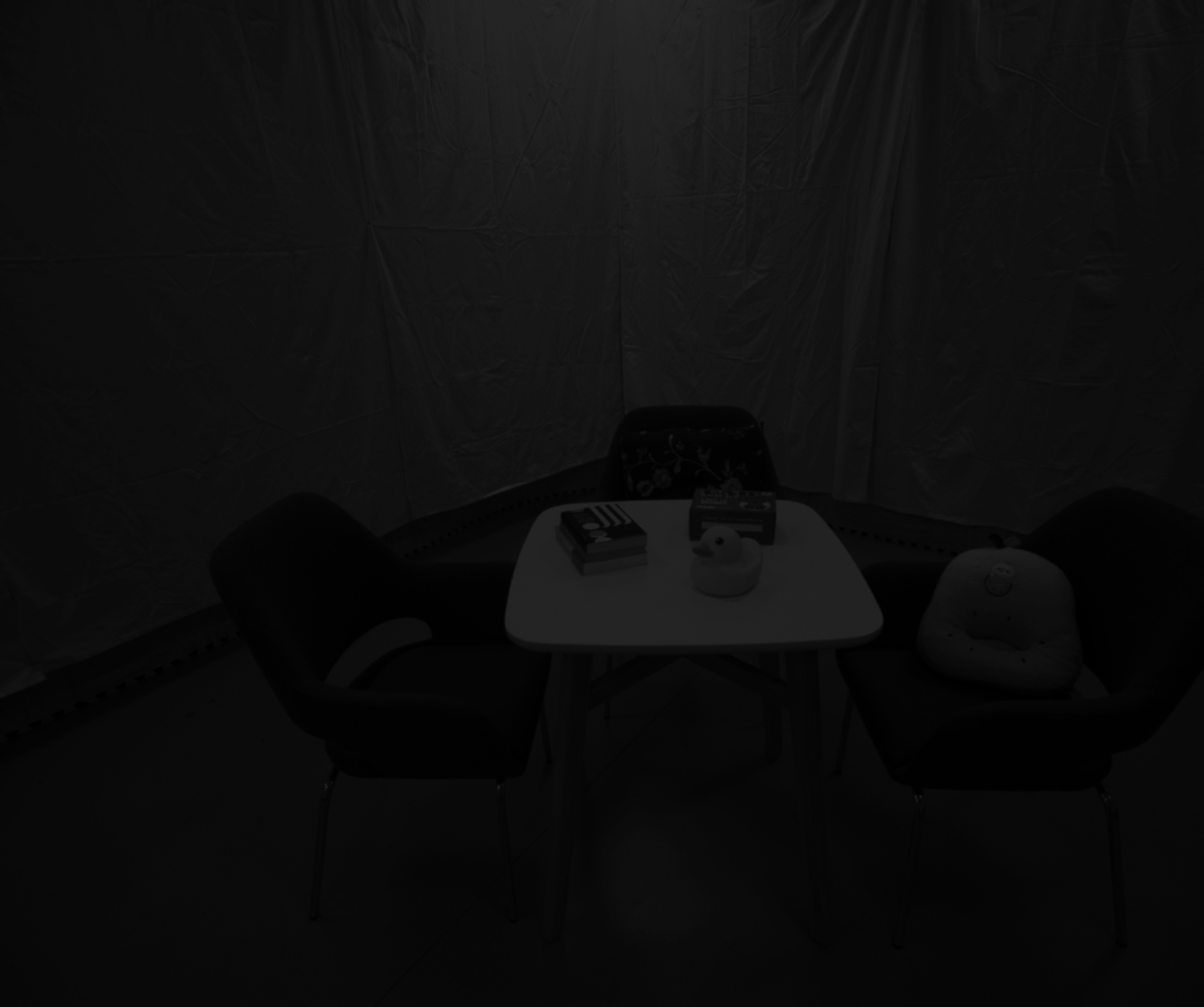}}
\subfigure[Overlay of events extracted]{\includegraphics[width=4.3cm]{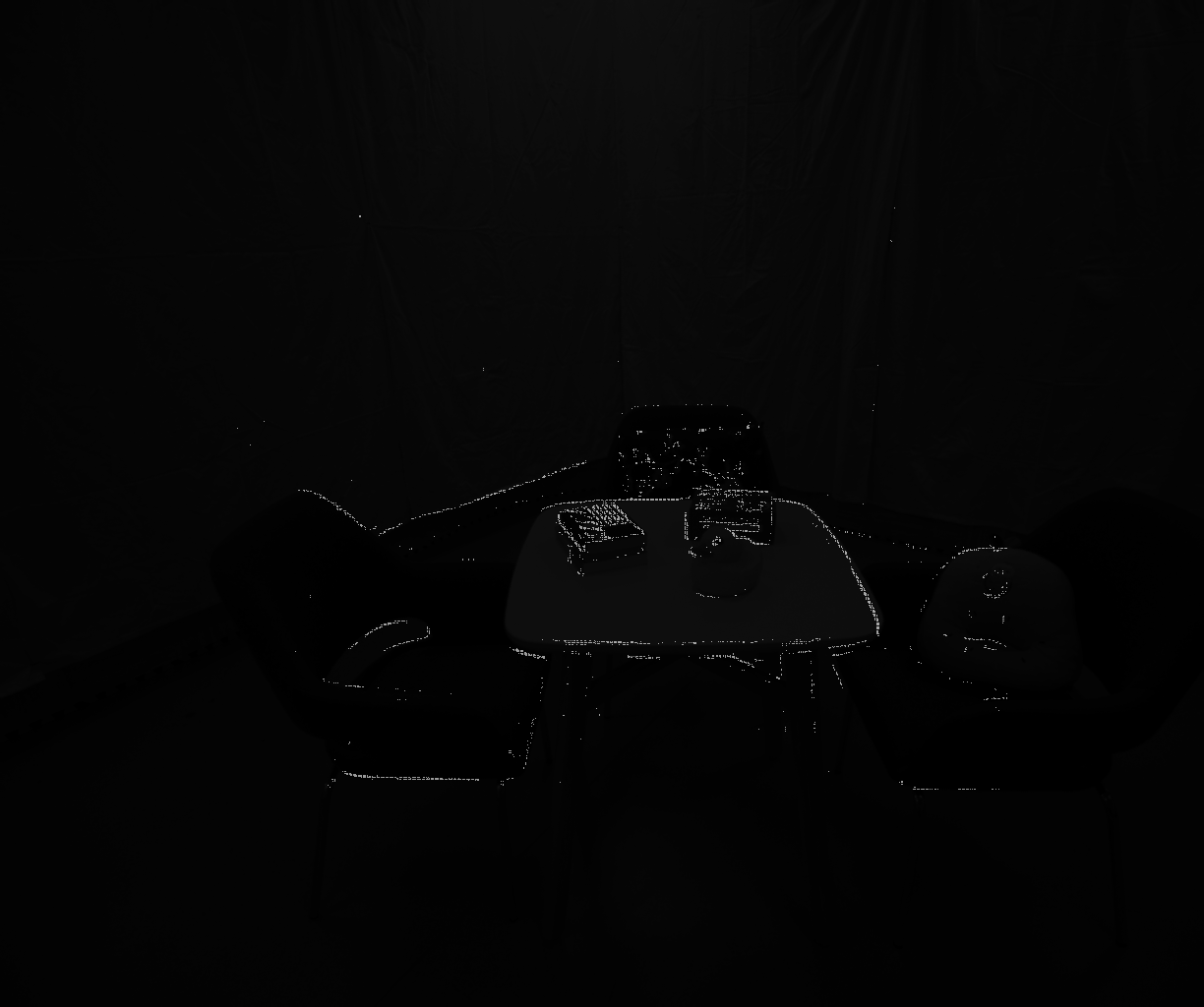}}
\caption{\label{scene in the dark}Visualization of the scene in the dark and an overlay of the extracted events.}
\end{figure}

\revise{
\subsubsection{Results for Different Illumination Conditions}
\label{results for different light conditions}
Some of the self-collected sequences explore the same scene but in different illumination conditions. As already outlined through the preceding results, in dark conditions, the frame-based \textbf{Canny-VT} fails while the event-based \textbf{Canny-EVT} continues to work. As can be observed qualitatively in Figure~\ref{scene in the dark}, while it is indeed hard to extract edges from images of the dark scene, events keep being triggered along the dominant appearance boundaries.}

\subsection{Discussion on the importance of cross-modality, polarity, and occlusion-handling}
\label{ablation study}

\revise{In order further underline the importance of generating the map from alternative sensors and then applying cross-modal tracking, we need to discuss the potential given by purely event-based methods. To start with, it is worth noting that the inadequacy of the quality of semi-dense maps generated solely from event data has already been pointed out in prior art~\cite{gallego2017event}. In order to further confirm this statement, we applied the state-of-the-art event-only method \textbf{EVO}\cite{rebecq2016evo} to the testing sequences. \textbf{EVO} utilizes a semi-dense tracker along with a ray-density-based structure extraction method called \textbf{EMVS}\cite{rebecq2018emvs}. Our finding is that not only does the method produce low-quality 3D representations, it fails to successfully track throughout practically all sequences, and thereby fails to produce a useful map at all. \textbf{EVO} alternates between tracking and mapping, and the framework depends on mild stop-and-go-like motion where the mapping module is being given sufficient time and data to locally converge before exploration can continue. Given the continuous exploration trajectories in the analyzed datasets, even the successful initialization of \textbf{EVO} proved to be highly challenging. In summary, the stability of current purely event-based monocular methods is insufficient to produce useful maps, and we believe that the use of the stereo approach \textbf{ESVO}~\cite{zhou2021event} is a good alternative to demonstrate the ability of current purely event-based methods.}

\revise{A final ablation study aims at proving the individual effectiveness of the polarity-aware registration and the occlusion handling. We denote \textbf{Canny-EVT-PR} and \textbf{Canny-EVT-OH} the versions on which only polarity aware registration or occlusion handling is active, respectively. The results for all methods are summarized in Tables \ref{tab:ablation study rotation} and \ref{tab:ablation study translation}. As can be observed, the addition of polarity-aware registration effectively improves the robustness of the system. For example, the system is unable to successfully process \textit{table aggressive} and \textit{desk normal} from the self-collected sequences if polarity-aware registration is not enabled. On the other hand, results on \textit{sofa normal} and \textit{robot normal} from the \textit{VECtor} datasets demonstrate the solid improvement in accuracy attained by the activation of occlusion handling. The best overall result is achieved by \textbf{Canny-EVT}.}

\section{Conclusion}\label{sec:conclusion}

In the past, edge-based, geometric semi-dense alignment has been demonstrated to be an efficient and highly accurate visual tracking solution. However, being dependent on regular vision-based sensors, the method has natural limitations in terms of motion dynamics and illumination conditions. In this work, we demonstrate how the approach represents an outstanding fit for event cameras to alleviate those problems, and we present solutions for both unknown and known environments. The first one generates local semi-dense point clouds online using a depth camera, while the second one gets away with only an event camera by tracking an existing global semi-dense point cloud of the environment. The proposed methods handle a large spectrum of challenging conditions, and find an excellent balance between performance and power or resource hungriness. Novel additions increase the method's ability to handle large-scale displacements, occlusions, and complicated texture. By releasing a unified framework containing all traditional and event camera based semi-dense solutions discussed in this work, we hope to make a valuable contribution to areas that require stable and versatile localization for intelligent mobile systems. \revisetwo{In our future work, we consider fusion with IMU signals in order to enhance the reliability of the present vision-only tracking system. Inertial pre-integration terms can provide precise initial poses and dynamics for non-linear optimization and optical flow prediction. Furthermore, inertial signals enable the regularization and parallel optimization of multiple successive frames. Another idea we currently explore is the development of an event-based place recognition module in order to support the initial localization of the camera.}

\begin{table*}[htbp]
  \centering
  \captionsetup{justification=centering}
  \caption{\revise{Ablation Study: Rotation RPE and ATE on self-collected, \textit{VECtor} and \textit{TUM} datasets for \textbf{Canny-EVT-{}-}, \textbf{Canny-EVT-PR}, \textbf{Canny-EVT-OH} and \textbf{Canny-EVT}. $\left[\mathbf{R}_\text{\upshape rpe}\text{\upshape : °/s},\mathbf{R}_\text{\upshape ate}\text{\upshape : °}\right]$}}
  \revise{
    \begin{tabular}{llccccccccccc}
    \toprule
          &       & \multicolumn{2}{c}{\textbf{Canny-EVT-{}-}} &       & \multicolumn{2}{c}{\textbf{Canny-EVT-PR}} &       & \multicolumn{2}{c}{\textbf{Canny-EVT-OH}} &       & \multicolumn{2}{c}{\textbf{Canny-EVT}} \\
\cmidrule{3-4}\cmidrule{6-7}\cmidrule{9-10}\cmidrule{12-13}    \textit{Dataset} & \textit{Sequence} & $\mathbf{R}_\text{rpe}$  & $\mathbf{R}_\text{ate}$  &       & $\mathbf{R}_\text{rpe}$  & $\mathbf{R}_\text{ate}$  &       & $\mathbf{R}_\text{rpe}$  & $\mathbf{R}_\text{ate}$  &       & $\mathbf{R}_\text{rpe}$  & $\mathbf{R}_\text{ate}$  \\
\cmidrule{1-4}\cmidrule{6-7}\cmidrule{9-10}\cmidrule{12-13}    \multirow{8}[2]{*}{\textit{Self-collected}} & \textit{table\_bright\_slow} & 1.13  & 10.3  &       & 0.98  & 4.13  &       & 1.14  & 6.65  &       & \textbf{0.98} & \textbf{4.10} \\
          & \textit{table\_bright\_mid} & 1.74  & 3.79  &       & 1.54  & 3.85  &       & \textbf{1.49} & 2.99  &       & 1.60  & \textbf{2.94} \\
          & \textit{table\_normal} & 0.79  & 3.96  &       & 0.85  & 3.83  &       & \textbf{0.76} & 3.82  &       & 0.82  & \textbf{3.81} \\
          & \textit{desk\_normal} & -     & -     &       & 1.00  & 2.36  &       & -     & -     &       & \textbf{1.00} & \textbf{2.35} \\
          & \textit{table\_bright\_fasl} & 1.65  & 3.22  &       & 1.61  & 3.08  &       & \textbf{1.58} & 3.16  &       & 1.59  & \textbf{3.07} \\
          & \textit{table\_aggressive} & -     & -     &       & 5.32  & 18.58 &       & -     & -     &       & \textbf{5.11} & \textbf{12.90} \\
          & \textit{sofa\_normal} & 9.87  & 16.44 &       & 1.01  & \textbf{2.27} &       & 1.18  & 2.55  &       & \textbf{1.00} & 2.31 \\
          & \textit{desk\_normal} & 0.95  & 2.43  &       & 0.92  & 2.38  &       & 0.87  & 2.38  &       & \textbf{0.85} & \textbf{2.35} \\
\cmidrule{1-4}\cmidrule{6-7}\cmidrule{9-10}\cmidrule{12-13}    \multirow{2}[2]{*}{\textit{VECtor}} & \textit{sofa\_normal} & 1.49  & 1.78  &       & \textbf{1.18} & 1.46  &       & 1.56  & 1.86  &       & 1.19  & \textbf{1.45} \\
          & \textit{sofa\_fast} & -     & -     &       & 2.74  & \textbf{3.10} &       & 2.67  & 4.50  &       & \textbf{2.57} & 3.14 \\
\cmidrule{1-4}\cmidrule{6-7}\cmidrule{9-10}\cmidrule{12-13}    \multirow{4}[2]{*}{\textit{TUM}} & \textit{fr2\_desk} & 0.014 & 0.015  &       & 0.012 & \textbf{0.011} &       & \textbf{0.011} & 0.012 &       & 0.013 & 0.012 \\
          & \textit{fr2\_rpy} & 0.004 & 0.002 &       & 0.003 & 0.002 &       & 0.003 & 0.002 &       & \textbf{0.003} & \textbf{0.002} \\
          & \textit{fr3\_long\_office} & 0.008  & 0.008  &       & 0.008  & 0.006  &       & \textbf{0.005} & \textbf{0.004} &       & 0.008  & 0.006 \\
          & \textit{fr1\_xyz} & 0.038  & 0.024  &       & 0.011 & 0.065 &       & 0.038  & 0.024  &       & \textbf{0.010} & \textbf{0.064} \\
    \bottomrule
    \end{tabular}%
    }
  \label{tab:ablation study rotation}%
\end{table*}%

\begin{table*}[htbp]
  \centering
  \captionsetup{justification=centering}
  \caption{\revise{Ablation Study: Translation RPE and ATE on self-collected, \textit{VECtor} and \textit{TUM} datasets for \textbf{Canny-EVT-{}-}, \textbf{Canny-EVT-PR}, \textbf{Canny-EVT-OH} and \textbf{Canny-EVT}. $\left[\mathbf{t}_\text{\upshape rpe}\text{\upshape : cm/s},\mathbf{t}_\text{\upshape ate}\text{\upshape : cm}\right]$}}
  \revise{
    \begin{tabular}{llccccccccccc}
    \toprule
          &       & \multicolumn{2}{c}{\textbf{Canny-EVT-{}-}} &       & \multicolumn{2}{c}{\textbf{Canny-EVT-PR}} &       & \multicolumn{2}{c}{\textbf{Canny-EVT-OH}} &       & \multicolumn{2}{c}{\textbf{Canny-EVT}} \\
\cmidrule{3-4}\cmidrule{6-7}\cmidrule{9-10}\cmidrule{12-13}    \textit{Dataset} & \textit{Sequence} & $\mathbf{t}_\text{rpe}$ & $\mathbf{t}_\text{ate}$ &       & $\mathbf{t}_\text{rpe}$ & $\mathbf{t}_\text{ate}$ &       & $\mathbf{t}_\text{rpe}$ & $\mathbf{t}_\text{ate}$ &       & $\mathbf{t}_\text{rpe}$ & $\mathbf{t}_\text{ate}$ \\
\cmidrule{1-4}\cmidrule{6-7}\cmidrule{9-10}\cmidrule{12-13}    \multirow{8}[2]{*}{\textit{Self-collected}} & \textit{table\_bright\_slow} & 2.72  & 3.01  &       & 1.01  & 1.13  &       & 2.28  & 2.43  &       & \textbf{1.01} & \textbf{1.13} \\
          & \textit{table\_bright\_mid} & 2.19  & 3.45  &       & 1.90  & 2.90  &       & 1.60  & 1.91  &       & \textbf{1.17} & \textbf{1.46} \\
          & \textit{table\_normal} & 1.55  & 1.19  &       & 1.38  & 1.03  &       & 1.52  & 1.17  &       & \textbf{1.31} & \textbf{0.99} \\
          & \textit{desk\_normal} & -     & -     &       & 0.99  & 0.83  &       & -     & -     &       & \textbf{0.99} & \textbf{0.83} \\
          & \textit{table\_bright\_fasl} & 1.84  & 2.50  &       & 1.74  & 2.39  &       & 1.74  & 2.43  &       & \textbf{1.71} & \textbf{2.39} \\
          & \textit{table\_aggressive} & -     & -     &       & 5.99  & 9.49  &       & -     & -     &       & \textbf{5.15} & \textbf{7.22} \\
          & \textit{sofa\_normal} & 49.70  & 48.74 &       & \textbf{0.88} & 0.77  &       & 1.29  & 1.16  &       & 0.89  & \textbf{0.76} \\
          & \textit{desk\_normal} & 1.11  & 0.82  &       & 0.96  & 0.76  &       & 0.99  & 0.77  &       & \textbf{0.70} & \textbf{0.72} \\
\cmidrule{1-4}\cmidrule{6-7}\cmidrule{9-10}\cmidrule{12-13}    \multirow{2}[2]{*}{\textit{VECtor}} & \textit{sofa\_normal} & 1.91  & 2.00  &       & 1.25  & 1.33  &       & 2.00  & 2.12  &       & \textbf{1.20} & \textbf{1.33} \\
          & \textit{sofa\_fast} & -     & -     &       & 1.56  & 1.43  &       & 2.46  & 1.87  &       & \textbf{1.50} & \textbf{1.43} \\
\cmidrule{1-4}\cmidrule{6-7}\cmidrule{9-10}\cmidrule{12-13}    \multirow{4}[2]{*}{\textit{TUM}} & \textit{fr2\_desk} & 0.92  & 1.39  &       & 0.72  & \textbf{0.73} &       & \textbf{0.69} & 1.07  &       & 0.74  & 0.79 \\
          & \textit{fr2\_rpy} & 0.36  & 10.39 &       & \textbf{0.25} & \textbf{4.59} &       & 0.41  & 9.23  &       & 0.27  & 6.69 \\
          & \textit{fr3\_long\_office} & 1.21  & 1.20  &       & 0.48  & 0.67  &       & \textbf{0.41} & \textbf{0.46} &       & 0.47  & 0.60 \\
          & \textit{fr1\_xyz} & 1.63  & 11.28 &       & 0.46 & \textbf{0.83} &       & 1.62  & 11.26 &       & \textbf{0.46} & 0.95 \\
    \bottomrule
    \end{tabular}%
    }
  \label{tab:ablation study translation}%
\end{table*}%

\section*{Acknowledgments}
The authors want to express their gratitude to the co-authors of the preceding work of this submission, Prof. Yi Zhou from Hunan University, and Prof. Hongdong Li from the Australian National University. Without their contributions, the publication of the edge-based tracking framework released open-source through this submission would not be possible. The authors would like to thank the fund support from the National Natural Science Foundation of China (62250610225) and Natural Science Foundation of Shanghai (22dz1201900, 22ZR1441300). We also want to acknowledge the generous support of and continued fruitful exchange with our project collaborators at Midea Robozone.

\newpage

\bibliographystyle{IEEEtran}
\bibliography{citation}

\begin{IEEEbiography}[{\includegraphics[width=1in,height=1.25in,clip,keepaspectratio]{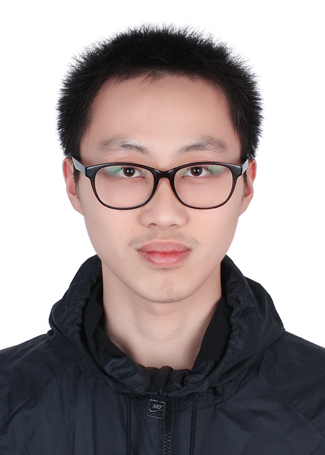}}]{Yi-Fan Zuo} Yi-Fan Zuo received the B.Sc degree in biological engineering from South China University of Technology, Guangzhou, China in 2017. He is currently pursuing the Ph.D. degree with the Beijing Institute
of Technology and a visiting student in  ShanghaiTech University. His research interests include visual odometry/simultaneous localization and mapping with multi-sensor and event cameras.
\end{IEEEbiography}

\begin{IEEEbiography}[{\includegraphics[width=1in,height=1.25in,clip,keepaspectratio]{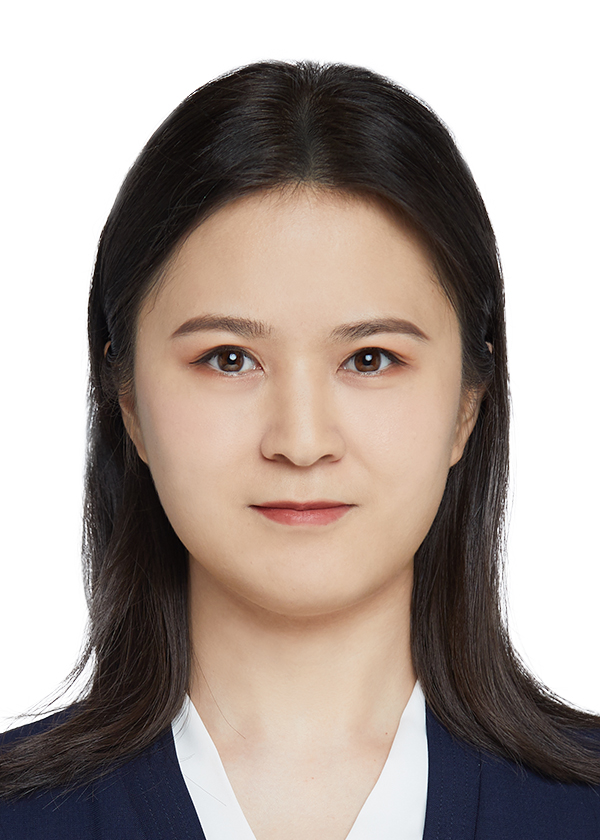}}]{Wanting Xu} Wanting Xu is currently a Ph.D. student in computer science at ShanghaiTech University, advised by Prof. Laurent Kneip. She received her B.S. degree in applied mathematics from Xinjiang University in 2018. Her research interests include visual SLAM and geometric computer vision, specifically focusing on the geometric solutions of pose estimation problems for traditional and event cameras.
\end{IEEEbiography}

\begin{IEEEbiography}[{\includegraphics[width=1in,height=1.25in,clip,keepaspectratio]{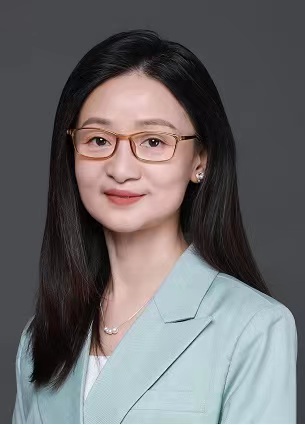}}]{Xia Wang} Dr. Xia Wang received the Ph.D. degree in Automation from the China University of Mining and Technology in 1999. She is currently an Associate Professor with the Beijing Institute of Technology, where she is also the Vice Dean of the Institute of Photoelectric Imaging and Information Engineering. Her current research interests include optoelectronic detection, spectrum analysis, and imaging technology.
\end{IEEEbiography}

\begin{IEEEbiography}[{\includegraphics[width=1in,height=1.25in,clip,keepaspectratio]{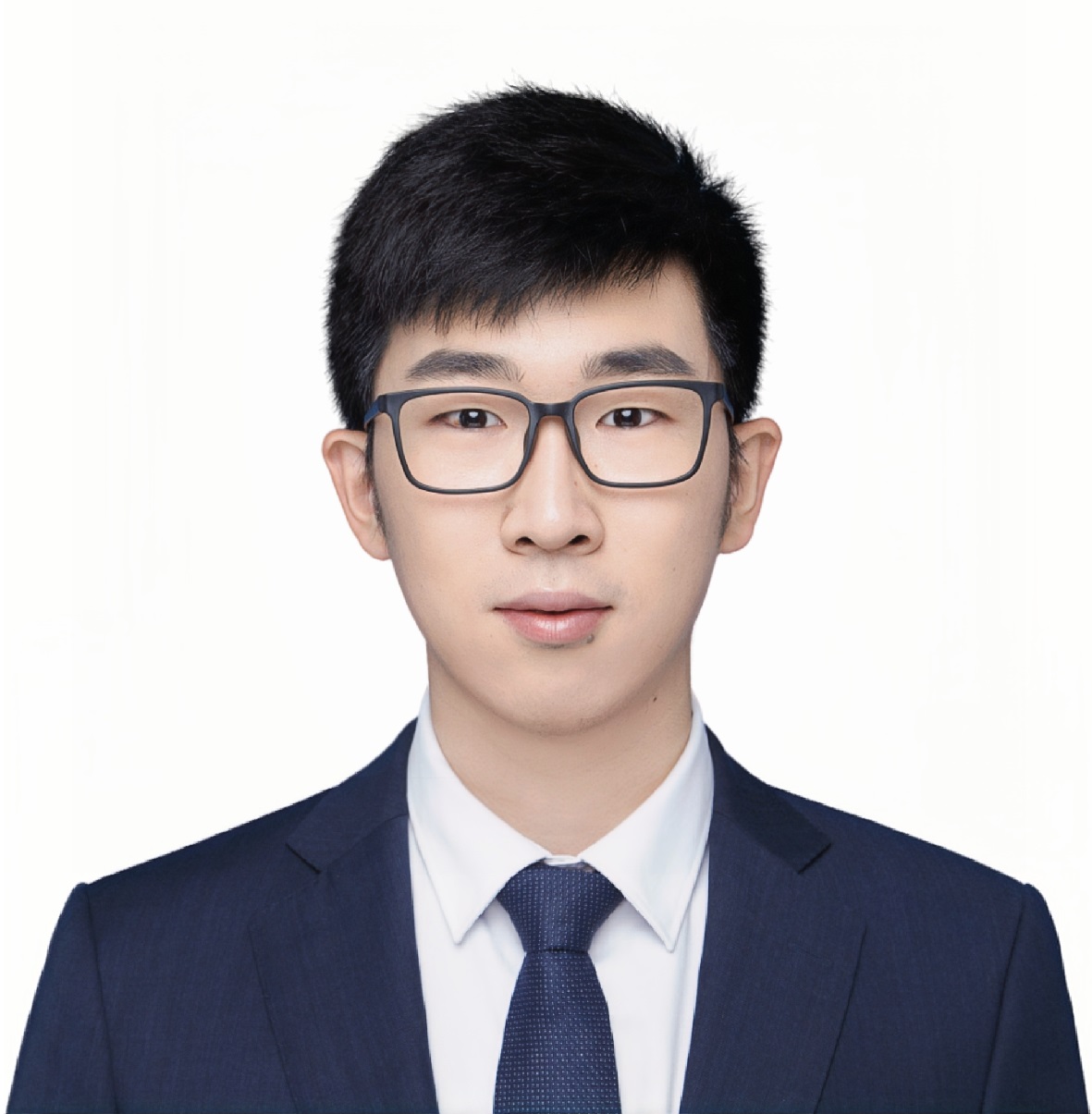}}]{Yifu Wang} Dr. Yifu Wang is currently a Postdoctoral Researcher at ShanghaiTech University. He was a Ph.D. student at Australian National University from 2016 to 2021, supervised by Prof. Laurent Kneip and Prof. Hongdong Li. Before that, he received his Bachelor's Degree in Engineering from Australian National University and Beijing Institute of Technology in 2015 and 2016 respectively. His research interests lie in robotic vision, particularly in visual odometry/SLAM for multi-camera systems and event-based-cameras.
\end{IEEEbiography}

\begin{IEEEbiography}[{\includegraphics[width=1in,height=1.25in,clip,keepaspectratio]{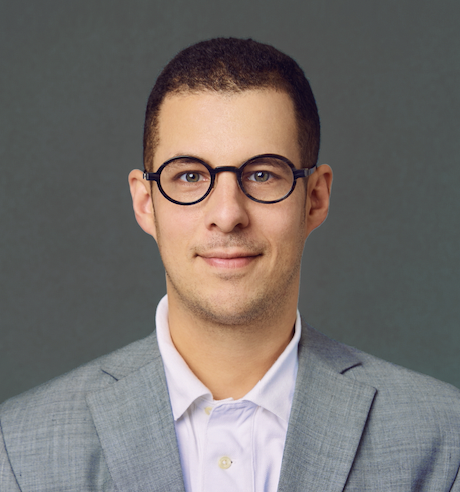}}]
{Laurent Kneip}
Dr Kneip owns a Dipl.-Ing. degree from the Friedrich-Alexander University Erlangen/Nürnberg, and a PhD degree from ETH Zurich, where he worked at the Autonomous Systems Lab. He is also a recipient of the ARC Discovery Early Career Researcher Award (DECRA) in 2015, and the Marr Prize (honourable mention) in 2017. Dr Kneip currently is tenured Associate Professor at ShanghaiTech University, where he founded and directs the Mobile Perception Laboratory. He is also the director of the ShanghaiTech Automation and Robotics center. Dr Kneip has countless publications in top robotics and computer vision venues, and continuous to research on enabling intelligent mobile systems to use vision for real-time 3D perception of the environment. Dr Kneip is the main author of OpenGV.
\end{IEEEbiography}

\vfill
\end{document}